\documentclass[lettersize,journal]{IEEEtran}
\usepackage{amsmath,amsfonts,amssymb}
\usepackage{algorithmic}
\usepackage{algorithm}
\usepackage{booktabs}
\usepackage{multicol}
\usepackage{multirow}
\usepackage{booktabs}
\usepackage{enumerate}
\usepackage{tcolorbox}
\usepackage{xcolor}
\usepackage{colortbl}
\usepackage{tabularray}
\usepackage{microtype}
\usepackage[utf8]{inputenc}
\usepackage[T1]{fontenc}
\UseTblrLibrary{booktabs}
\usepackage{tabularx}
\usepackage{hhline}
\usepackage{bbding}
\usepackage{tikz}
\usepackage{paralist}
\usepackage{array}
\usepackage[caption=false,font=scriptsize,labelfont=sf,textfont=sf]{subfig}
\usepackage{textcomp}
\usepackage{stfloats}
\usepackage{soul}
\usepackage{url}
\usepackage{verbatim}
\usepackage{graphicx}
\usepackage{hhline}
\usepackage[colorlinks,
linkcolor=red,
anchorcolor=black,
citecolor=red]{hyperref}
\usepackage{orcidlink}
\usepackage{overpic}
\def\ie{\emph{i.e.}}
\def\eg{\emph{e.g.}}

\begin{document}

\title{Decoupled Motion Representation Learning for Moving Infrared Small Target Detection}

\author{Guoyi~Zhang,~Peiwen~Wu,~Han~Wang,~Xiangpeng~Xu,~and~Xiaohu~Zhang
	\thanks{Guoyi~Zhang, Peiwen~Wu, Han~Wang, Xiangpeng~Xu, and Xiaohu~Zhang are with the School of Aeronautics and Astronautics, Sun Yat-sen University, Shenzhen 518107, Guangdong, China. (Corresponding authors: \emph{Han Wang, Xiangpeng Xu and Xiaohu Zhang}, email:\href{mailto:zhanggy57@mail2.sysu.edu.cn}{zhanggy57@mail2.sysu.edu.cn})}
}

\markboth{Journal of \LaTeX\ Class Files,~Vol.~14, No.~8, August~2021}%
{Zhang \MakeLowercase{\textit{et al.}}: Decoupled Motion Representation Learning for Moving Infrared Small Target Detection}


\maketitle
\begin{abstract}
Infrared small target detection in dynamic scenes remains challenging due to the highly coupled motions among targets, imaging platforms, and dynamic backgrounds. Existing multi-frame methods usually perform implicit temporal modeling, where coherent background dynamics dominate motion correspondence learning, leading to an inherent trade‑off between detection and false alarms. In this work, we observe that background motions exhibit strong global coherence, whereas small targets mainly correspond to sparse local motion anomalies. Moreover, many false-alarm responses maintain high consistency with globally coherent motion patterns, indicating that they mainly originate from coherent background dynamics rather than genuine target motions.
Based on these observations, we propose a decoupled motion representation learning framework for moving infrared small target detection. Specifically, an explicit motion branch is introduced to model globally coherent motion dynamics using pretrained optical flow priors, together with a structure-preserving self-supervised adaptation strategy for infrared motion correspondence learning. Meanwhile, an implicit motion branch based on deformable feature alignment is designed to capture target-sensitive local motion anomalies under coherent motion guidance. Furthermore, a coherent-motion-guided local anomaly reasoning module is proposed to identify and suppress coherent-motion-induced false responses during localized motion modeling.
Extensive experiments on two challenging infrared small target detection benchmarks demonstrate that the proposed method consistently outperforms existing state-of-the-art approaches, particularly in dynamic scenes with complex motions, while maintaining favorable inference efficiency.
\end{abstract}


\section{Introduction} \label{Sec:Intro}
\begin{figure}[!t]
	\centering
	\includegraphics[width=\linewidth]{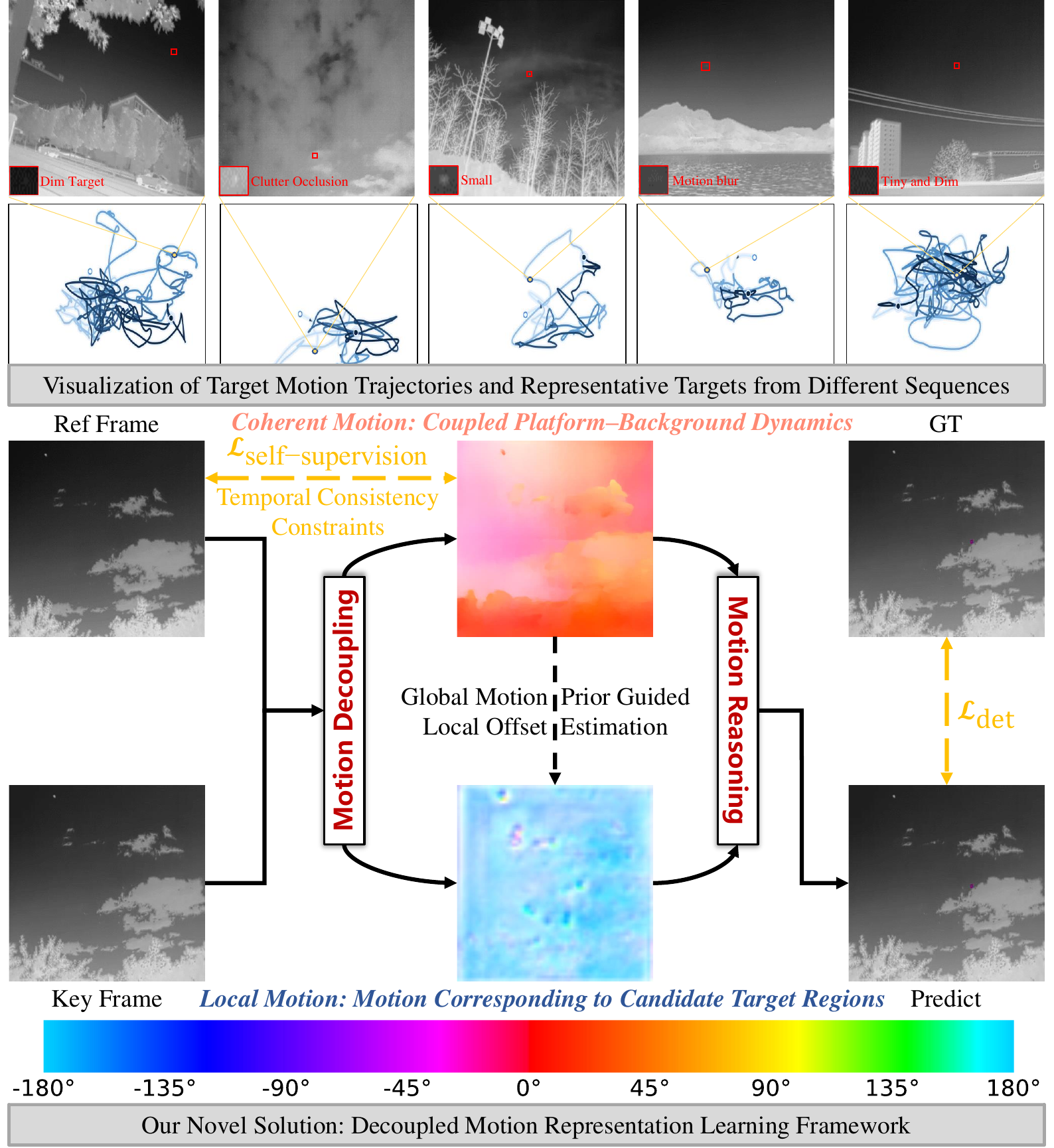}
	\caption{Motion information is critical for infrared dim small target detection in cluttered scenes, yet effective exploitation remains challenging due to highly coupled and complex motion patterns. Our \textbf{key observation} is that background-induced false alarms usually exhibit strong global motion consistency, whereas target motions manifest as localized anomalies. Based on this observation, we propose a motion-decoupled representation framework that jointly models global motion consistency and local motion anomalies for robust target detection.}
	\label{fig:teaser}
\end{figure}

\IEEEPARstart{I}{nfrared} small target detection has been widely applied in long-range surveillance, early warning, and aerial tracking systems \cite{UIUNet} due to its ability to perceive dim and long-distance targets under complex environments \cite{11080263}. Although remarkable progress has been achieved in static scenes, reliable detection in dynamic scenarios remains highly challenging \cite{DNANet,11278553}. In practical applications, imaging platforms, dynamic backgrounds, and moving targets often exhibit highly coupled motions under camera motion and scene variations \cite{gao2026mist}. As illustrated in Fig.~\ref{fig:teaser}, such coupled interactions result in highly entangled and irregular motion trajectories. Under such highly entangled motion patterns, coherent background dynamics tend to dominate temporal correspondence learning, causing background-induced motion responses to be progressively reinforced while sparse target motions are gradually submerged, thereby inducing severe false alarms.

\begin{figure}[!t]
	\centering
	\includegraphics[width=\linewidth]{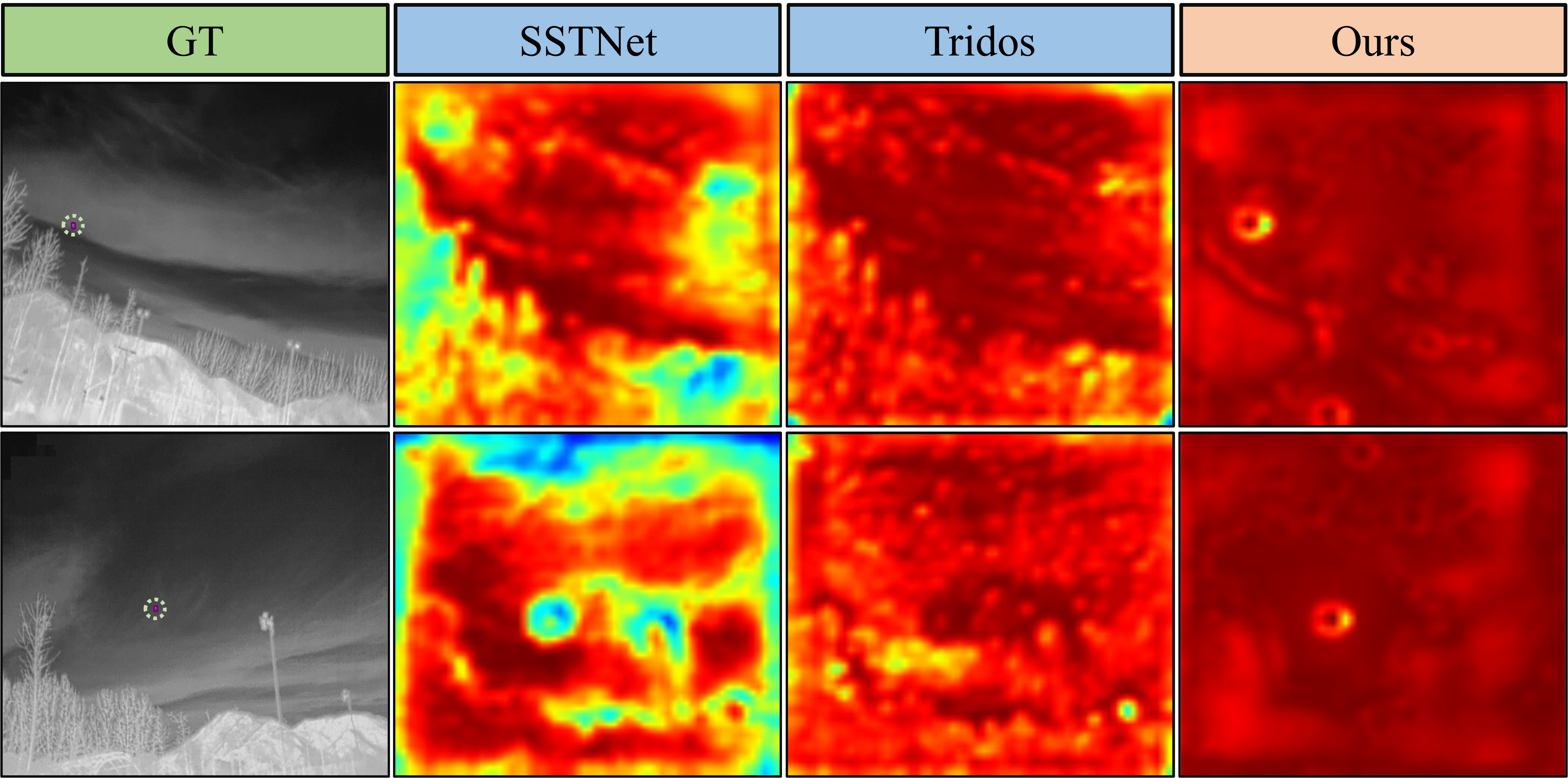}
	\caption{Visualization of motion ambiguity induced by implicit temporal modeling. During temporal aggregation, coherent background responses are progressively reinforced, while sparse target-induced local motions become gradually submerged, leading to severe feature confusion and false-alarm accumulation. PCA is applied for feature visualization.}
	\label{fig:Degrading}
\end{figure}

Existing multi-frame approaches fundamentally rely on temporal correspondence aggregation to enhance motion perception across frames \cite{chen2024sstnet,duan2024triple,chen2025motion}. However, such implicit temporal modeling intrinsically favors globally coherent motions, since background dynamics usually produce spatially continuous and temporally stable correspondence responses over large regions. In contrast, infrared small targets usually produce sparse, weak, and locally irregular motion responses due to their tiny scale and low signal-to-noise characteristics. Consequently, as shown in Fig.~\ref{fig:Degrading},  localized target motions are easily submerged during temporal aggregation, while coherent background responses become progressively reinforced, leading to severe motion ambiguity and false-alarm accumulation.

To address the above challenges, we observe that background motions usually exhibit strong global coherence, whereas small targets mainly correspond to sparse localized motion anomalies. More importantly, many false alarms remain highly consistent with coherent background motions, causing implicit temporal aggregation to progressively reinforce background responses while suppressing weak target motions. Based on this observation, we propose a decoupled motion representation learning framework for moving infrared small target detection, which explicitly separates globally coherent motion reasoning from localized motion anomaly perception. Specifically, an explicit branch models coherent motion dynamics using optical flow priors, while an implicit branch captures target-sensitive local motion anomalies via deformable alignment. Furthermore, coherent motion priors are utilized to guide local motion reasoning, thereby suppressing background-induced false responses and enhancing genuine target motion representations.

The main contributions of this work are summarized as follows:
\begin{itemize}
	\item We propose a decoupled motion representation learning framework for moving infrared small target detection, which explicitly separates globally coherent motion dynamics and localized target motion anomalies to alleviate severe motion entanglement in temporal modeling.
	\item We design a dual-branch motion modeling paradigm, where an explicit branch captures globally coherent motion dynamics using pretrained optical flow priors, while an implicit branch performs deformable local alignment to model target-sensitive local motion anomalies under the guidance of coherent motion representations.
	\item We introduce a structure-preserving self-supervised optical flow adaptation strategy, which transfers pretrained motion correspondence priors to infrared scenarios while preserving motion-sensitive structural consistency.
	\item We propose a coherent-motion-guided conditional reasoning mechanism, which exploits the consistency between explicit coherent motions and implicit local motion responses to suppress background-induced false alarms and enhance target-sensitive local motion representations.
\end{itemize}

The remainder of this paper is organized as follows: In
Section \ref{Section:Related_Work}, related works are briefly reviewed. In Section \ref{Section:ProposedMethod},
we present the proposed method in detail. In Section \ref{Section:Experiment}, the
experimental results are given and discussed. Conclusions are
drawn in Section \ref{Section:Conclusion}.

\section{Related Work} \label{Section:Related_Work}
\subsection{Single-Frame Infrared Small Target Detection} Early studies on infrared small target detection predominantly adopted single-frame segmentation paradigms, leveraging multi-scale contextual modeling \cite{DNANet,UIUNet}, long-range feature interactions \cite{liu2023infrared}, shape-biased representations \cite{CSRNet}, wavelet-based downsampling \cite{11278553}, and multi-task mutual learning \cite{11080263} to enhance detection performance. While effective in relatively simple scenes, these approaches are fundamentally constrained by their reliance on appearance-based cues \cite{li2023direction}. In long-range surveillance scenarios, true targets and clutter often exhibit nearly indistinguishable spatial characteristics, with the primary discriminative information encoded in their temporal motion behaviors \cite{Duan_2026_CVPR}. Consequently, the absence of temporal modeling limits the robustness of single-frame methods and leads to degraded performance in complex dynamic environments.
\subsection{Multi-Frame Infrared Small Target Detection} Multi-frame methods formulate infrared small target detection as a spatio-temporal feature matching problem by exploiting temporal cues across consecutive frames \cite{gong2021temporal}, which provide discriminative information beyond instantaneous appearance. Existing approaches can be broadly categorized into explicit motion modeling, implicit motion modeling, and vision-language multimodal methods. Explicit methods \cite{10989627,li2023direction,11145128,Duan_2026_CVPR} directly estimate target trajectories or motion patterns, offering strong interpretability; however, the irregular motions of small targets and interference from dynamic backgrounds make accurate motion modeling difficult, and their performance is sensitive to limited training data in small-scale datasets \cite{10989627}. Implicit methods \cite{zhu2024tmp,10409231,duan2024triple,chen2024sstnet,peng2025moving,10824834,deng2026learning,gao2026mist} learn temporal representations without predefined motion priors, offering greater flexibility but suffering from limited interpretability and weak motion supervision. More recently, vision-language methods \cite{chen2025motion,duan2026sevil} introduce language priors to facilitate temporal reasoning and improve performance, but require additional annotation efforts and remain constrained by the difficulty of describing complex motion dynamics in natural language. Despite their differences, these methods focus on modeling target motion and implicitly assume inter-frame target consistency. This assumption can be severely degraded in complex dynamic scenes, where coupled large-scale motions of the imaging platform and time-varying background introduce substantial motion blur and appearance distortions.
\subsection{Motivation}
Unlike existing methods that directly model target motion, we observe that most false alarms exhibit motion patterns consistent with the dominant coherent motion induced by the imaging platform and time-varying background. This observation suggests that the key challenge is not merely motion modeling itself, but distinguishing local motion anomalies from scene-level coherent motion. Therefore, explicitly disentangling scene-level coherent motion from local motion anomalies provides a promising direction for robust target localization in complex dynamic environments.

\section{Methodology} \label{Section:ProposedMethod}
\subsection{Problem Formulation} \label{Section:ProblemFormulation}
Given an infrared sequence $\boldsymbol{\mathcal{X}}=\{\mathbf{X}_t\}_{t=1}^{T}$, the observed motion field $\boldsymbol{\mathcal{M}}_t$ is jointly affected by camera motion, dynamic backgrounds, and target motion. Existing temporal modeling methods generally learn motion representations via $\mathbf{F}_t=\mathcal{F}(\boldsymbol{\mathcal{M}}_t)$, where $\mathcal{F}(\cdot)$ denotes implicit temporal aggregation.
However, the observed motion is dominated by globally coherent background dynamics rather than independently observable target motions. Specifically,
\begin{equation}
	\boldsymbol{\mathcal{M}}_t=\Phi(\boldsymbol{\mathcal{M}}_t^{g},\boldsymbol{\mathcal{M}}_t^{l}),
\end{equation}
where $\boldsymbol{\mathcal{M}}_t^{g}$ and $\boldsymbol{\mathcal{M}}_t^{l}$ denote globally coherent motion and localized motion anomaly, respectively, and $\Phi(\cdot)$ represents nonlinear motion entanglement. Typically, coherent motions satisfy $\nabla \boldsymbol{\mathcal{M}}_t^{g}\approx0$, while localized target motions exhibit sparse spatial discontinuities, \ie, $\|\nabla\boldsymbol{\mathcal{ M}}_t^{l}\|\gg\|\nabla \boldsymbol{\mathcal{M}}_t^{g}\|$.
Under such nonlinear coupling, temporal correspondence learning is easily dominated by coherent background motions, resulting in severe motion ambiguity and false-alarm responses.

Therefore, the core objective is to recover target-sensitive localized motion representations:
\begin{equation}
	\mathbf{F}_t^{l}=\Psi(\boldsymbol{\mathcal{M}}_t^{l}),
\end{equation}
from the entangled motion field $\boldsymbol{\mathcal{M}}_t=\Phi(\boldsymbol{\mathcal{M}}_t^{g},\boldsymbol{\mathcal{M}}_t^{l})$,
while suppressing globally coherent background interference induced by $\boldsymbol{\mathcal{M}}_t^{g}$. However, since $\Phi(\cdot)$ is highly nonlinear, directly separating the two motion components through linear decomposition is intractable. Therefore, the key challenge lies in disentangling localized target motion anomalies from coherent background dynamics under nonlinear temporal coupling.
\subsection{Overview}
\begin{figure*}[!t]
	\centering
	\includegraphics[width=\linewidth]{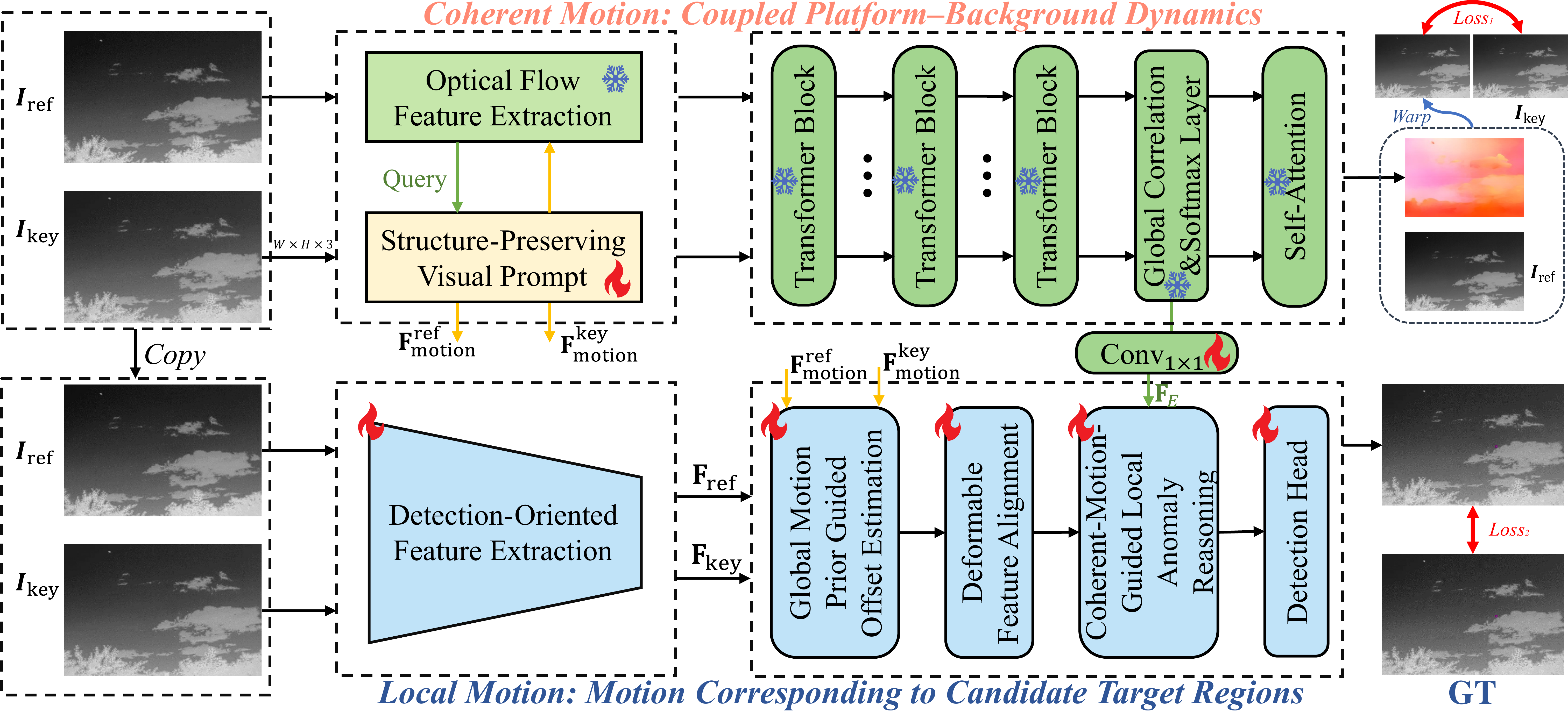}
	\caption{Overview of the proposed DMR framework. Motivated by the observation that coherent background dynamics dominate temporal correspondence learning while target motions mainly manifest as localized anomalies, the proposed framework explicitly disentangles globally coherent motion and local motion anomalies through complementary explicit and implicit motion branches. The resulting coherent-motion priors further guide local anomaly reasoning to suppress coherent-motion-consistent false responses and enhance target-sensitive motion representations. Here, \includegraphics[height=1em]{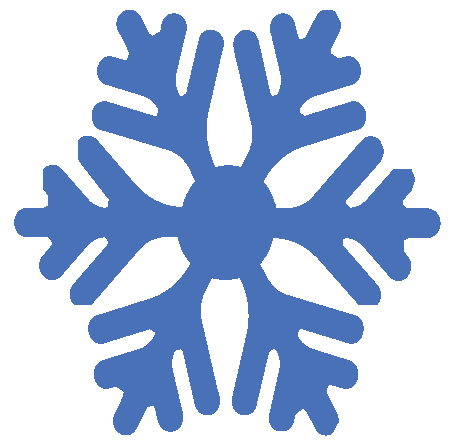} and \includegraphics[height=1em]{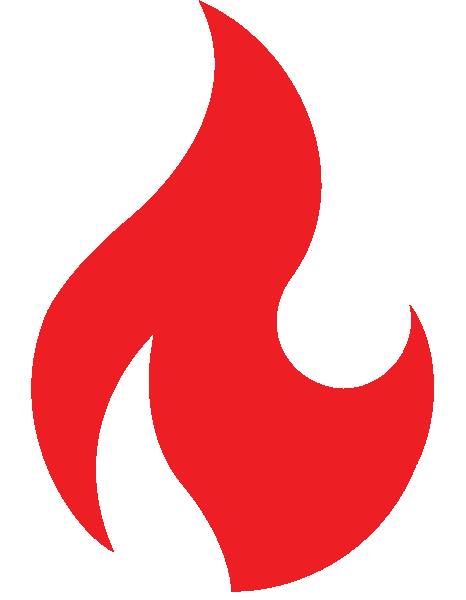} represent frozen and learnable modules, respectively. Moreover, $\textit{Loss}_1$ and $\textit{Loss}_2$ denote the temporal consistency self-supervised loss and detection loss, respectively.}
	\label{fig:arch}
\end{figure*}
As shown in Fig. \ref{fig:arch}, we propose a decoupled motion representation learning framework to disentangle globally coherent motion dynamics and localized motion anomalies from the entangled motion field $\boldsymbol{\mathcal{M}}_t=\Phi(\boldsymbol{\mathcal{M}}_t^{g},\boldsymbol{\mathcal{M}}_t^{l})$. Specifically, the explicit motion branch models globally coherent motion representations $\mathbf{F}_t^{g}=\Psi_g(\boldsymbol{\mathcal{M}}_t^{g})$ via optical flow priors, while the implicit motion branch aims to recover target-sensitive localized motion representations $\mathbf{F}_t^{l}=\Psi_l(\boldsymbol{\mathcal{M}}_t \mid \mathbf{F}_t^{g})$ through deformable local motion modeling conditioned on coherent-motion priors. Furthermore, a coherent-motion-guided local anomaly reasoning module is introduced to perform coherent-motion-guided refinement:
$\hat{\mathbf{F}}_t^{l}=\mathcal{R}(\mathbf{F}_t^{l}\mid\mathbf{F}_t^{g})$,
thereby suppressing coherent-motion-induced false responses and enhancing target-sensitive local motion anomalies.
\subsection{Explicit Coherent Motion Branch}
The explicit motion branch aims to explicitly model globally coherent motion dynamics $\mathbf{F}_t^{g}=\Psi_g(\boldsymbol{\mathcal{M}}_t^{g})$ from entangled temporal representations. To explicitly characterize such dominant coherent dynamics, we introduce a pretrained optical flow model \cite{xu2023unifying} to extract stable globally coherent motion priors for subsequent localized motion reasoning.
\subsubsection{Structure-Preserving Visual Prompt Adaptation}
Directly transferring pretrained optical flow models to infrared imagery suffers from severe modality discrepancy, while full fine-tuning easily destroys pretrained motion correspondence priors. To preserve motion-sensitive temporal correspondence during infrared adaptation, we introduce a structure-preserving prompt adaptation strategy.

\textbf{\textit{Multi-Scale Prompt Extraction}}.
A lightweight pyramid extractor is employed to capture multi-scale structure-sensitive representations for infrared motion modeling. Specifically, the extractor adopts a three-stage hierarchical structure, where each stage consists of a $3\times3$ convolution followed by a stride-2 convolution for progressive spatial downsampling, producing multi-scale features $[\mathbf{F}_{1},\mathbf{F}_{2},\mathbf{F}_{3}]$:
\begin{equation}
	\mathbf{X}
	=
	\mathrm{Conv}_{1\times1}
	(
	[\mathrm{PS}(\mathbf{F}_{1}),
	\mathrm{PS}(\mathbf{F}_{2}),
	\mathbf{F}_{3}]
	),
\end{equation}
where $\mathrm{PS}(\cdot)$ denotes the PixelShuffle operation.

\textbf{\textit{Structure-Preserving Phase Enhancement}}.
Compared with amplitude information that is sensitive to thermal fluctuations, phase information mainly preserves structural correspondence and contour continuity, making it more robust for motion modeling under low-SNR infrared scenarios. Therefore, a lightweight phase-aware enhancement strategy is introduced to strengthen structure-sensitive motion representations.
Given an input feature $\mathbf{X}\in\mathbb{R}^{C\times H\times W}$, we first split it into:
\begin{equation}
	\mathbf{X}
	=
	[\mathbf{X}_p,\mathbf{X}_r],
\end{equation}
where $\mathbf{X}_p$ is used for phase-aware modulation and $\mathbf{X}_r$ preserves the remaining semantic representations.
To alleviate smooth background interference and noisy high-frequency responses in infrared imagery, a learnable frequency modulation strategy is introduced:
\begin{equation}
	\mathbf{Z}_p
	=
	\mathcal{F}^{-1}
	(
	\mathcal{F}(\mathbf{X}_p)
	\odot
	\mathbf{L}
	)
	+
	\mathbf{X}_p,
\end{equation}
where $\mathcal{F}(\cdot)$ and $\mathcal{F}^{-1}(\cdot)$ denote the Fourier transform and inverse Fourier transform, respectively, $\mathbf{L}$ is a learnable frequency modulation matrix, and $\odot$ denotes element-wise multiplication.
The phase spectrum is extracted and adaptively refined:
\begin{equation}
	\mathbf{P}
	=
	\operatorname{Phase}
	(
	\mathcal{F}(\mathbf{Z}_p)
	), \quad \hat{\mathbf{P}}
	=
	\mathbf{P}
	+
	\phi(\mathbf{P}),
\end{equation}
where $\phi(\cdot)$ denotes a lightweight MLP.
The refined phase representation is reconstructed into the spatial domain to generate a structure-aware modulation map:
\begin{equation}
	\mathbf{A}
	=
	\sigma
	\left(
	\operatorname{Re}
	\left(
	\mathcal{F}^{-1}
	(
	e^{j\hat{\mathbf{P}}}
	)
	\right)
	\right),
\end{equation}
where $\sigma(\cdot)$ denotes the sigmoid activation function.
The phase-aware representation is then recalibrated as:
\begin{equation}
	\mathbf{X}_p'
	=
	\mathbf{X}_p
	\odot
	(1+\alpha\mathbf{A}),
\end{equation}
where $\alpha$ is a learnable scalar. Finally, the enhanced representation is fused with the remaining semantic features:
\begin{equation}
	\mathbf{Y}
	=
	\mathbf{X}
	+
	f_{fuse}
	(
	[\mathbf{X}_p',\mathbf{X}_r]
	),
	\label{Eq:VP}
\end{equation}
where $f_{fuse}(\cdot)$ models inter-channel interactions.

\textbf{\textit{Cross-Attention Interaction}}.
Direct additive fusion may overwrite pretrained motion correspondence priors. Therefore, a cross-attention interaction mechanism is introduced to enable adaptive interaction between infrared-aware prompts and frozen motion priors. Given the prompt feature $\mathbf{Y}$ and frozen motion representation $\mathbf{X}$ extracted by the optical flow backbone, the interaction process is formulated as:
\begin{equation}
	\mathbf{Q}
	=
	\mathbf{W}_q\mathbf{Y},
	\quad
	\mathbf{K}
	=
	\mathbf{W}_k\mathbf{X},
	\quad
	\mathbf{V}
	=
	\mathbf{W}_v\mathbf{X},
\end{equation}
\begin{equation}
	\mathbf{Z}
	=
	\mathrm{Softmax}
	\left(
	\frac{\mathbf{Q}\mathbf{K}^{\top}}{\sqrt{d}}
	\right)
	\mathbf{V}
	+
	\mathbf{X},
\end{equation}
where $\mathbf{W}_q$, $\mathbf{W}_k$, and $\mathbf{W}_v$ are learnable projection matrices, and $d$ denotes the channel dimension. Unlike direct fusion, cross-attention preserves pretrained temporal correspondence while enabling adaptive infrared-aware interaction.
\subsubsection{Self-supervised Temporal Correspondence Learning}
Detection supervision alone cannot reliably preserve temporal correspondence during infrared adaptation, causing the motion branch to gradually drift toward appearance-dominant representations. Once motion-sensitive correspondence is corrupted, coherent background dynamics can no longer be reliably characterized, further aggravating temporal ambiguity and false-alarm accumulation during subsequent local motion modeling. To preserve stable coherent-motion priors, self-supervised temporal consistency constraints \cite{10154030} are introduced to regularize motion-sensitive correspondence learning during infrared adaptation.

During training, the optical flow backbone remains frozen, while only the prompt extractor and lightweight interaction modules are optimized. Given an image pair $(I_1,I_2)$, the model predicts bidirectional multi-scale optical flows
$\{\mathbf{F}_{1\rightarrow2}^{(s)},\mathbf{F}_{2\rightarrow1}^{(s)}\}_{s=1}^{S}$.
The photometric consistency loss at scale $s$ is defined as:
\begin{equation}
	\begin{aligned}
		\mathcal{L}_{ph}^{(s)}
		&=
		\left\|
		\mathbf{M}^{(s)}
		\odot
		\left(
		\mathcal{W}
		(
		I_2^{(s)},
		\mathbf{F}_{1\rightarrow2}^{(s)}
		)
		-
		I_1^{(s)}
		\right)
		\right\|_1 \\
		&\quad+
		\lambda_{ssim}
		\mathcal{L}_{ssim}
		\left(
		\mathcal{W}
		(
		I_2^{(s)},
		\mathbf{F}_{1\rightarrow2}^{(s)}
		),
		I_1^{(s)}
		\right),
	\end{aligned}
\end{equation}
where $\mathcal{W}(\cdot)$ denotes differentiable warping and $\mathbf{M}^{(s)}$ denotes the occlusion mask estimated by forward-backward consistency.
Although infrared imagery suffers from weak textures and thermal fluctuations, short-term temporal correspondence still provides stable motion constraints. Therefore, the self-supervised objective regularizes temporal consistency during infrared adaptation and prevents pretrained motion representations from drifting toward appearance-dominant features. The overall self-supervised loss is $\mathcal{L}_{self}
=
\sum_{s=1}^{S}
\mathcal{L}_{ph}^{(s)}$.
\subsubsection{Discussion}
\begin{figure}[!t]
	\centering
	\includegraphics[width=\linewidth]{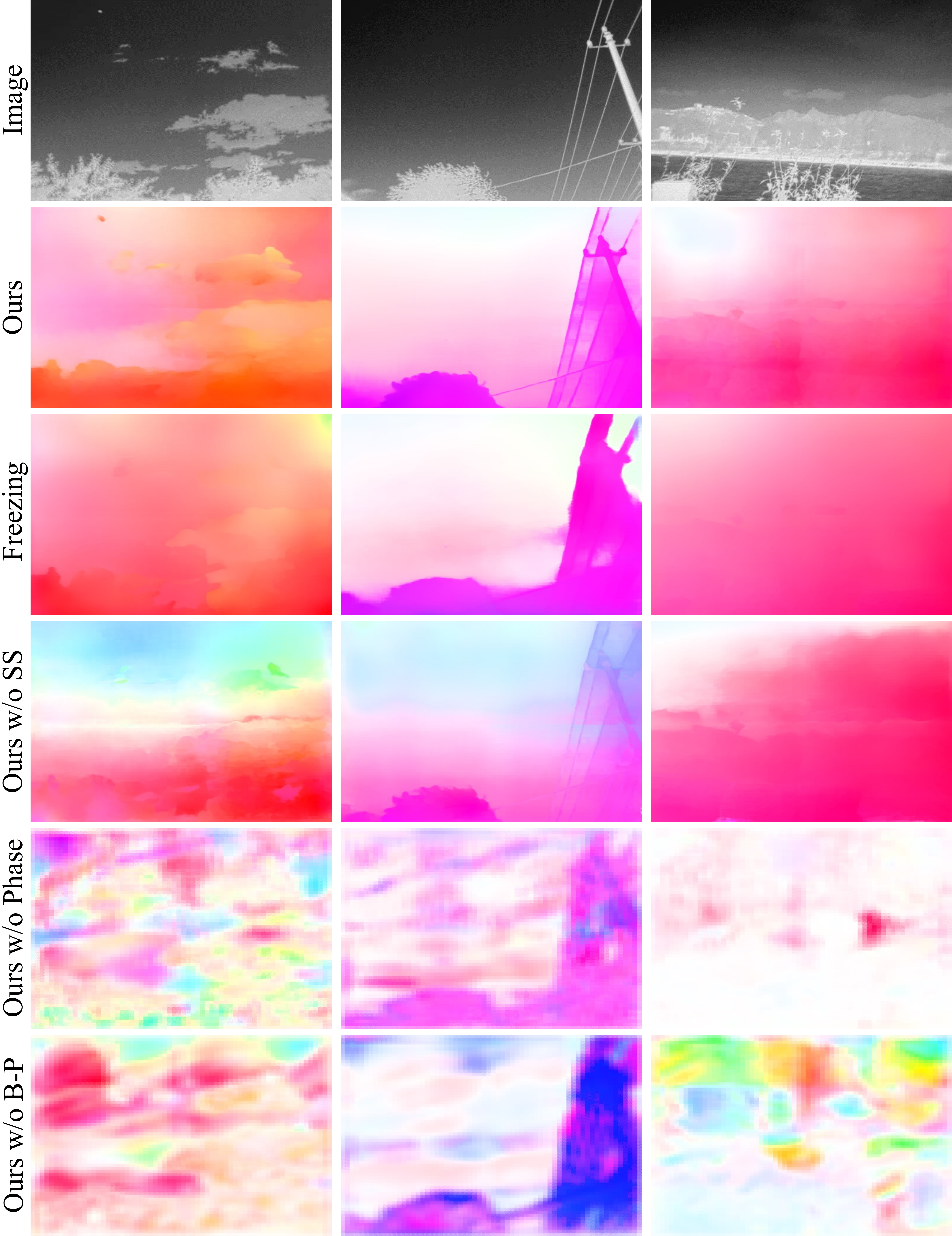}
	\caption{Visual comparison of optical flow estimation under different designs. When the optical flow model is fully frozen, the estimated flow maps are coarse due to the modality discrepancy between visible-domain motion priors and infrared imagery. Introducing the phase-aware prompt improves structural perception and produces finer motion details; however, noticeable local abnormal responses still remain without self-supervised temporal constraints. Removing the phase-aware enhancement severely degrades flow estimation quality, indicating that structural correspondence is critical for reliable infrared motion modeling. Similarly, removing the band-pass filtering introduces substantial noisy responses, since infrared imagery contains strong high-frequency interference. Overall, the proposed strategy effectively preserves structural motion correspondence while suppressing local abnormal responses. Here, ``SS'' denotes the self-supervised strategy, and ``B-P'' denotes the band-pass filtering module.
	}
	\label{fig:op}
\end{figure}
\begin{table}[!t]
	\renewcommand\arraystretch{1.}
	\centering
	\caption{Comparison of Parameter-Efficient Fine-Tuning Methods on IRDST Dataset.}
	\label{Tab:PEFT}
	\setlength{\tabcolsep}{4.0pt}
	\resizebox{0.48\textwidth}{!}{\begin{tabular}{l|cccc}
			\noalign{\hrule height 1pt}
			\textbf{Methods}                    & $\textbf{mAP}_\textbf{50}(\%)$$\uparrow$    & $\textbf{Pr}(\%)$$\uparrow$    & $\textbf{Re}(\%)$$\uparrow$            & \textbf{F1}(\%)$\uparrow$                                      \\ \noalign{\hrule height 1pt}
			Freezing & \underline{73.33} & \textbf{90.90} & 81.32 & \underline{85.85} \\
			Full-Tuning & 71.30 & 88.29 & 81.67 & 84.85 \\
			\noalign{\hrule height 1pt}
			VPT \cite{jia2022visual}  \tiny\textcolor{red}{\textit{(ECCV'22)}} & 56.00 & 75.10 & 75.84 & 75.47 \\
			ConvLoRA \cite{DBLP:conf/iclr/ZhongTHFY24} \tiny\textcolor{red}{\textit{(ICLR'24)}} & 62.48 & 79.25 & 79.85 & 79.55 \\
			LMSA \cite{gao2024multi} \tiny\textcolor{red}{\textit{(MM'24)}} & 68.21 & 85.42 & 80.19 & 82.72 \\
			ViT-CoMer \cite{Xia_2024_CVPR} \tiny\textcolor{red}{\textit{(CVPR'24)}} & 71.23 & 87.25 & \underline{82.04} & 84.56 \\
			MONA \cite{yin20255} \tiny\textcolor{red}{\textit{(CVPR'25)}} & 69.90 & 86.59 & 81.27 & 83.85 \\
			FBNM \cite{10844993}  \tiny\textcolor{red}{\textit{(TMM'25)}} & 63.68 & 82.75 & 77.25 & 79.91 \\
			EMIP \cite{10989627} \tiny\textcolor{red}{\textit{(TIP'25)}} & 68.00 & 86.32 & 79.95 & 83.01 \\
			EVPv2 \cite{11197268} \tiny\textcolor{red}{\textit{(TPAMI'26)}} & 63.95 & 83.14 & 77.70 & 80.33 \\
			HFCP \cite{jia2026vit} \tiny\textcolor{red}{\textit{(TIP'26)}} & 69.17 & 85.58 & 81.32 & 83.39 \\
			\noalign{\hrule height 1pt}
			Ours  & \textbf{79.13} & \underline{87.53} & \textbf{91.23}  &  \textbf{89.34}  \\ \noalign{\hrule height 1pt}
	\end{tabular}}
\end{table}
To further validate the proposed adaptation strategy, different parameter-efficient fine-tuning (PEFT) methods are compared in Tab.~\ref{Tab:PEFT}. Since infrared optical flow annotations are unavailable, the motion branch is optimized only under detection supervision. Under this weak-supervision setting, full fine-tuning severely degrades performance because pretrained motion correspondence priors are gradually destroyed during infrared adaptation. Existing PEFT methods also show limited effectiveness, as direct backbone adaptation tends to disturb pretrained temporal correspondence representations and becomes sensitive to the heavy noise characteristics of infrared imagery.

In contrast, the proposed method achieves the best performance by explicitly preserving motion-sensitive correspondence during infrared adaptation. Specifically, the self-supervised constraint regularizes temporal consistency, while the phase-aware enhancement suppresses noisy responses and strengthens structure-sensitive motion representations. As illustrated in Fig.~\ref{fig:op}, the explicit motion branch effectively captures globally coherent motion dynamics, but still struggles to model the sparse localized motions of infrared small targets, which further motivates the introduction of the implicit motion branch for localized motion anomaly modeling.
\subsection{Implicit Localized Motion Anomaly Branch}
While globally coherent motion modeling helps characterize dominant background dynamics $\mathbf{F}_t^{g}=\Psi_g(\boldsymbol{\mathcal{M}}_t^{g})$, localized target-induced motion anomalies are easily submerged during temporal aggregation due to their sparse and weak responses. Consequently, directly recovering target-sensitive local motions from entangled temporal representations $\boldsymbol{\mathcal{M}}_t$ remains highly challenging. To address this issue, the implicit motion branch is introduced to recover residual localized motion anomalies $\mathbf{F}_t^{l}=\Psi_l(\boldsymbol{\mathcal{M}}_t \mid \mathbf{F}_t^{g})$ under the guidance of coherent-motion priors provided by the explicit branch. Specifically, the motion representations extracted from the explicit branch are incorporated into offset estimation to provide motion-aware structural cues under weak-texture infrared scenes.
\subsubsection{Detection-Oriented Feature Extraction}
Following prior work \cite{chen2024sstnet,duan2024triple}, key-frame and reference-frame features are extracted using the YOLOX backbone and FPN \cite{yolox2021}. Considering that infrared small targets mainly appear as weak low-level structures in high-resolution feature maps, only the highest-resolution FPN features ($\mathbf{F}_{key}$ and $\mathbf{F}_{ref}$) are utilized for subsequent motion alignment.
\subsubsection{Global Motion Prior Guided Offset Estimation}
Accurate localized motion recovery heavily depends on stable deformable offset estimation. However, in infrared dynamic scenes, sparse target responses and weak structural textures make local motion estimation highly unstable. Since the explicit branch preserves temporally coherent motion correspondence, the resulting coherent-motion priors provide reliable structural guidance for suppressing background-induced motion ambiguity during deformable offset prediction.
For the key-frame and reference-frame features, the motion-enhanced representations are formulated as:
\begin{equation}
	\mathbf{F}_{offset}^{key}
	=
	\mathrm{Conv}_{1\times1}
	(
	[\mathbf{F}_{key},\mathbf{F}_{motion}^{key}]
	),
\end{equation}
\begin{equation}
	\mathbf{F}_{offset}^{ref}
	=
	\mathrm{Conv}_{1\times1}
	(
	[\mathbf{F}_{ref},\mathbf{F}_{motion}^{ref}]
	),
\end{equation}
where $\mathbf{F}_{motion}^{key}$ and $\mathbf{F}_{motion}^{ref}$ denote the motion-sensitive representations extracted from the explicit motion branch (Eq.~\ref{Eq:VP}).
The offset estimation network takes the concatenated motion-aware representation as input:
\begin{equation}
	\mathbf{X}_0
	=
	\mathrm{Conv}_{1\times1}
	(
	[\mathbf{F}_{offset}^{key},
	\mathbf{F}_{offset}^{ref}]
	).
\end{equation}
Since localized target motions may exhibit different spatial ranges, we further employ a multi-receptive-field refinement strategy to progressively aggregate motion-aware contextual representations:
\begin{equation}
	\mathbf{X}_i =
	\mathcal{F}_i \big( \mathbf{X}_{i-1} + \mathbf{S}_{i-1} \big)
	+ \mathbf{X}_{i-1}, \quad i=1,\dots,4,
\end{equation}
where $\mathbf{S}_{i-1} = \sum_{k=0}^{i-1} \mathbf{X}_k$ aggregates hierarchical residual states, and $\mathcal{F}_i(\cdot)$ denotes a dilated convolution block with dilation rate $d_i$. Different receptive fields enable the model to perceive localized motion patterns with varying spatial extents.
The deformable offsets are finally predicted as:
\begin{equation}
	\Delta \mathbf{P}
	=
	\mathrm{Conv}_{3\times3}
	(
	\mathbf{X}_4
	),
\end{equation}
where $\Delta \mathbf{P}\in\mathbb{R}^{2K\times H\times W}$ denotes the learned spatial offsets for deformable convolution, and $K$ is the number of deformable sampling locations.
\subsubsection{Deformable Feature Alignment}
The predicted offsets are used to spatially align the reference-frame feature toward the key-frame feature via deformable convolution:
\begin{equation}
	\mathbf{F}_{align}(\mathbf{p}_0)
	=
	\sum_{k=1}^{K}
	w_k
	\cdot
	\mathbf{F}_{ref}
	(
	\mathbf{p}_0
	+
	\mathbf{p}_k
	+
	\Delta\mathbf{p}_k
	),
\end{equation}
where $\mathbf{p}_0$ denotes the reference spatial location, $\mathbf{p}_k$ are the regular convolution grid points, $\Delta\mathbf{p}_k$ are the learned offsets, and $w_k$ are the convolution weights. Such adaptive sampling enables spatially flexible localized motion alignment under complex background interference.
To further enhance target-sensitive motion responses, the aligned feature is utilized to adaptively modulate the original key-frame representation:
\begin{equation}
	\mathbf{F}_{out}
	=
	\mathbf{F}_{align}
	\odot
	\mathbf{F}_{key}
	+
	\mathbf{F}_{key},
\end{equation}
where $\mathbf{F}_{out}$ denotes the localized motion representation produced by the implicit motion branch. Since $\mathbf{F}_{align}$ encodes temporally aligned local motion cues, the modulation operation enhances target-sensitive local motion anomalies while suppressing inconsistent background variations.
\subsubsection{Discussion}
\begin{figure}[!t]
	\centering
	\includegraphics[width=\linewidth]{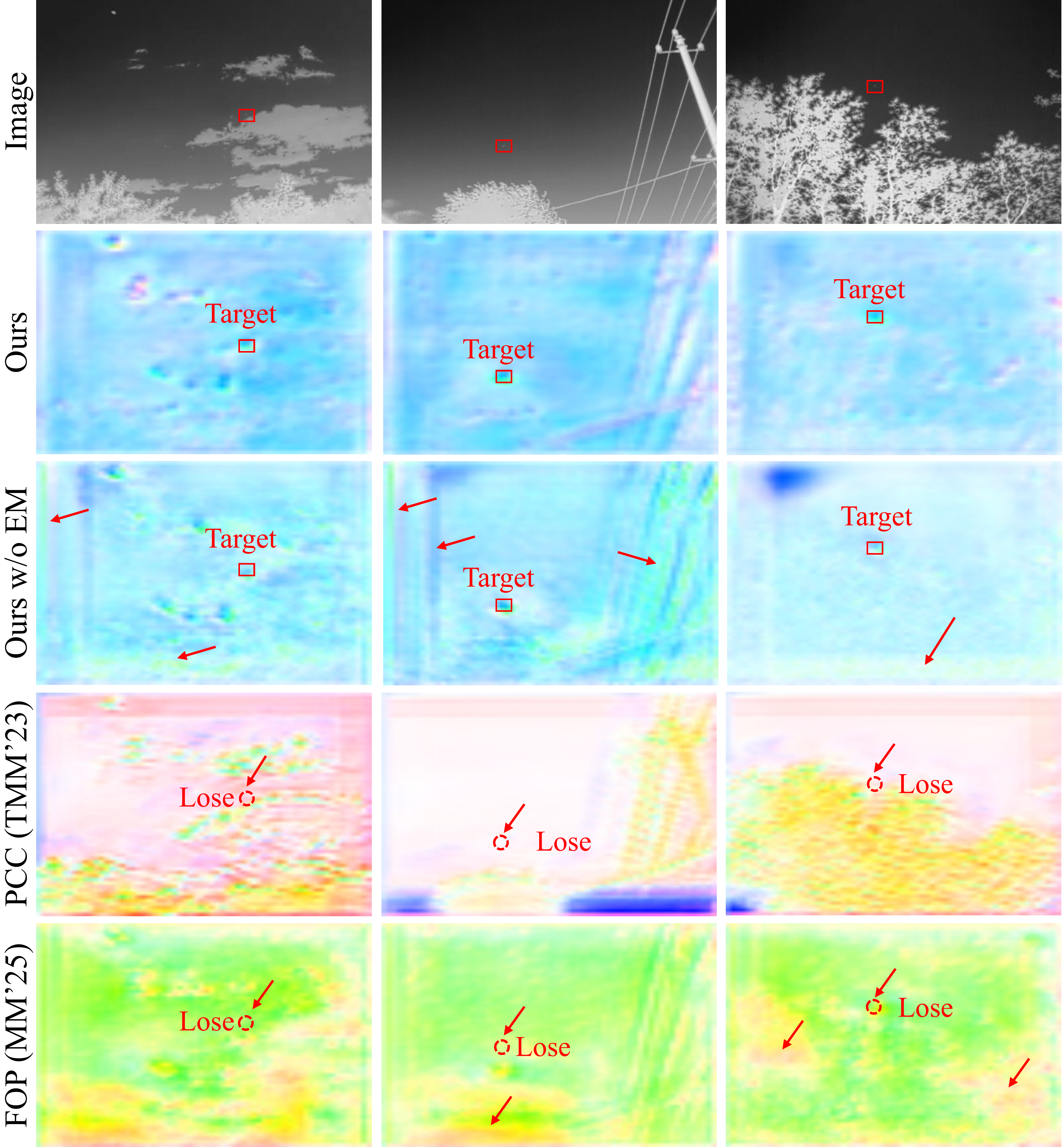}
	\caption{Visualization comparison of deformable convolution offsets under different designs. Benefiting from the guidance of the explicit motion branch, the estimated offsets become more stable and exhibit enhanced sensitivity to localized target motion patterns. In contrast, optical flow estimation is inherently less effective at modeling the subtle motions of infrared small targets, thereby causing the flow-guided strategy to suffer from the loss of target-specific motion details. Here, ``EM'' denotes the guidance introduced by the explicit motion branch, while ``FOP'' refers to the use of optical flow estimation results to guide offset prediction.}
	\label{fig:offset}
\end{figure}
\begin{table}[!t]
	\renewcommand\arraystretch{1.}
	\centering
	\caption{Comparison of Different Deformable Offset Estimation Methods on IRDST Dataset. ``EM'' denotes the guidance introduced by the explicit motion branch}
	\label{Tab:Offset}
	\setlength{\tabcolsep}{4.0pt}
	\resizebox{0.48\textwidth}{!}{\begin{tabular}{l|cccc}
			\noalign{\hrule height 1pt}
			\textbf{Methods}                    & $\textbf{mAP}_\textbf{50}(\%)$$\uparrow$    & $\textbf{Pr}(\%)$$\uparrow$    & $\textbf{Re}(\%)$$\uparrow$            & \textbf{F1}(\%)$\uparrow$                                      \\ \noalign{\hrule height 1pt}
			PCC \cite{9919404} \tiny\textcolor{red}{\textit{(TMM'23)}} & 71.21 & 87.25 & 82.04 & 84.56 \\
		     FOP \cite{10.1145/3746027.3755718} \tiny\textcolor{red}{\textit{(MM'25)}} & 72.80 & 89.02 & 82.43 & 85.60 \\
		     Ours w/o EM & 77.80 & \textbf{90.89} & 86.52 & 88.65 \\
		     \noalign{\hrule height 1pt}
			Ours  & \textbf{79.13} & \underline{87.53} & \textbf{91.23}  & \textbf{89.34}  \\ \noalign{\hrule height 1pt}
	\end{tabular}}
\end{table}
The reliability of deformable feature alignment largely depends on stable offset estimation \cite{chan2021understanding}. However, directly constraining deformable offsets using optical flow estimation is often unreliable for infrared weak-target scenarios, since optical flow estimation is easily dominated by globally coherent background dynamics under low-SNR conditions, causing the learned offsets to become sensitive to background interference and appearance fluctuations. As reported in Tab. \ref{Tab:Offset}, such flow-consistent offset supervision leads to noticeable performance degradation compared with the proposed motion-prior-guided strategy.

Unlike previous methods that directly constrain deformable offsets using optical flow estimation results, the proposed self-supervised optical flow adaptation strategy enhances the structural perception ability of visual prompts during temporal correspondence learning. Such structure-aware representations provide more reliable guidance for offset estimation, improving its stability and accuracy. By introducing these visual prompts into deformable alignment, the implicit branch can better focus on residual localized motion anomaly of tiny targets. As shown in Fig. \ref{fig:offset}, the resulting offsets become more stable and target-aware, which further motivates the subsequent coherent-motion-guided local anomaly reasoning.

\subsection{Coherent-Motion-Guided Local Anomaly Reasoning}
According to the formulation in Sec. \ref{Section:ProblemFormulation}, the observed motion field $\boldsymbol{\mathcal{M}}_t=\Phi(\boldsymbol{\mathcal{M}}_t^{g},\boldsymbol{\mathcal{M}}_t^{l})$ is dominated by globally coherent background dynamics, causing localized target motions to become heavily entangled with coherent background motions during temporal correspondence learning. Consequently, although the implicit motion branch attempts to recover target-sensitive localized motion representations $\mathbf{F}_t^{l}=\Psi_l(\boldsymbol{\mathcal{M}}_t \mid \mathbf{F}_t^{g})$ under coherent-motion priors provided by the explicit branch, the recovered local motion representations may still remain conditionally correlated with globally coherent motions due to the highly nonlinear motion entanglement $\Phi(\cdot)$. As a result, coherent-motion-induced local responses can still be preserved in $\mathbf{F}_t^{l}$, thereby introducing substantial false-alarm interference. Therefore, beyond localized motion recovery, it is further necessary to perform conditional coherent-motion reasoning to progressively refine the recovered local motion representations under explicit coherent-motion priors:
\begin{equation}
	\hat{\mathbf{F}}_t^{l}
	=
	\mathcal{R}
	(
	\Psi_l
	(
	\boldsymbol{\mathcal{M}}_t
	\mid
	\mathbf{F}_t^{g}
	)
	\mid
	\mathbf{F}_t^{g}
	),
\end{equation}
where $\mathcal{R}(\cdot)$ denotes the conditional coherent-motion refinement function.

\begin{figure}[!t]
	\centering
	\includegraphics[width=\linewidth]{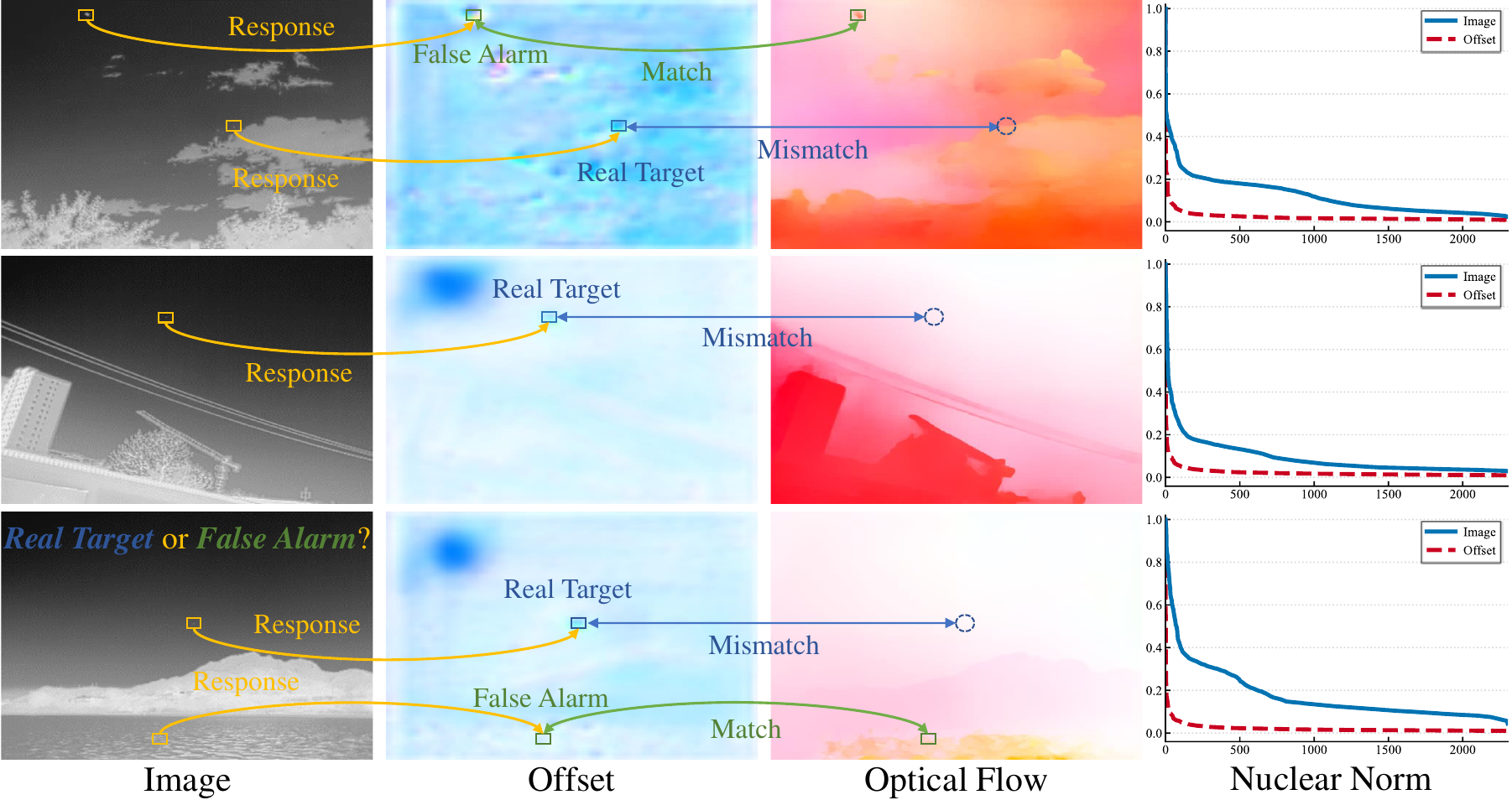}
	\caption{Optical flow, deformable offset visualization, and normalized nuclear norm curves. The false-alarm local motions from the implicit branch are highly consistent with the explicit branch, indicating a coherent background origin; the offset features exhibit significantly lower nuclear norm, confirming the global coherence of false alarms and inspiring feature refinement via autocorrelation-based discrimination. These observations motivate the proposed coherent-motion-guided local anomaly reasoning module.}
	\label{fig:co}
\end{figure}

As illustrated in Fig.~\ref{fig:co}, many false-alarm local motion responses exhibit high consistency with globally coherent motions, indicating that such responses mainly originate from coherent background dynamics rather than genuine target-induced motion anomalies. Motivated by this observation, we propose an explicit--implicit conditional motion reasoning framework, where the explicit branch models globally coherent motion dynamics $\boldsymbol{\mathcal{M}}_t^{g}$ and the implicit branch focuses on localized motion anomalies $\boldsymbol{\mathcal{M}}_t^{l}$.
Instead of directly aggregating explicit and implicit motion features, the proposed reasoning mechanism utilizes coherent motion representations to condition localized motion responses. Responses highly consistent with coherent background motion are suppressed, while localized inconsistent responses are preserved to enhance target-sensitive motion representations. Therefore, the explicit branch is utilized to identify coherent-motion-consistent false responses during localized motion reasoning:
\begin{equation}
	\hat{\boldsymbol{\mathcal{M}}}_t^{l}
	=
	\boldsymbol{\mathcal{M}}_t^{l}
	+
	\mathcal{F}
	(
	\boldsymbol{\mathcal{M}}_t^{l}
	\mid
	\boldsymbol{\mathcal{M}}_t^{g}
	),
\end{equation}
where $\mathcal{F}(\cdot)$ denotes the conditional refinement function guided by coherent-motion priors.
To instantiate the above conditional formulation, we first enhance motion-consistent local responses within the implicit branch via self-attention aggregation, where motion-consistent local structures are selectively aggregated through feature interaction. The refined implicit representation is subsequently utilized as the query to retrieve coherent-motion priors from the explicit branch via cross-attention, thereby enabling conditional refinement under coherent-motion guidance.
Specifically, self-attention is first performed on the implicit motion feature:
\begin{equation}
	\tilde{\mathbf{F}}_I
	=
	\mathbf{F}_I
	+
	\mathrm{Softmax}
	\left(
	\frac{\mathbf{Q}_I\mathbf{K}_I^\top}{\sqrt{d}}
	\right)
	\mathbf{V}_I,
\end{equation}
with
\begin{equation}
	\mathbf{Q}_I
	=
	\mathbf{W}_Q^I\mathbf{F}_I,
	\quad
	\mathbf{K}_I
	=
	\mathbf{W}_K^I\mathbf{F}_I,
	\quad
	\mathbf{V}_I
	=
	\mathbf{W}_V^I\mathbf{F}_I.
\end{equation}

The refined implicit representation subsequently interacts with the explicit motion feature through cross-attention to realize the conditional refinement process $\hat{\mathbf{F}}_t^{l}=\mathcal{R}(\Psi_l(\boldsymbol{\mathcal{M}}_t \mid \mathbf{F}_t^{g}) \mid \mathbf{F}_t^{g})$:
\begin{equation}
	\hat{\mathbf{F}}_I
	=
	\tilde{\mathbf{F}}_I
	+
	\mathrm{Softmax}
	\left(
	\frac{\mathbf{Q}\mathbf{K}^\top}{\sqrt{d}}
	\right)
	\mathbf{V},
\end{equation}
with
\begin{equation}
	\mathbf{Q}
	=
	\mathbf{W}_Q\tilde{\mathbf{F}}_I,
	\quad
	\mathbf{K}
	=
	\mathbf{W}_K\mathbf{F}_E,
	\quad
	\mathbf{V}
	=
	\mathbf{W}_V\mathbf{F}_E.
\end{equation}
Here, the implicit branch serves as the query to identify motion-consistent local responses, while the explicit branch provides globally coherent motion priors for conditional motion reasoning. Since coherent background motions usually exhibit strong global consistency, false-alarm local responses tend to maintain high motion correlation with the explicit branch. In contrast, genuine target-induced local motion anomalies generally exhibit relatively weak coherent-motion consistency. Therefore, the proposed hierarchical conditional reasoning mechanism enables the implicit branch to distinguish target-sensitive local motion anomalies from coherent-motion-induced false responses.

The conditionally refined representation is further aggregated through a feed-forward transformation to instantiate the target-sensitive localized motion recovery function $\mathbf{F}_t^{l}=\Psi(\boldsymbol{\mathcal{M}}_t^{l})$:
\begin{equation}
	\mathbf{F}_{out}
	=
	\hat{\mathbf{F}}_I
	+
	\mathrm{FFN}
	(
	\mathrm{LN}
	(
	\hat{\mathbf{F}}_I
	)
	).
\end{equation}
By explicitly modeling globally coherent motion dynamics and conditionally refining localized motion anomalies, the proposed framework effectively alleviates nonlinear motion entanglement during temporal correspondence learning, thereby enabling complementary modeling of coherent background dynamics and target-sensitive localized motion anomaly.

\subsubsection{Discussion}
\begin{figure}[!t]
	\centering
	\includegraphics[width=\linewidth]{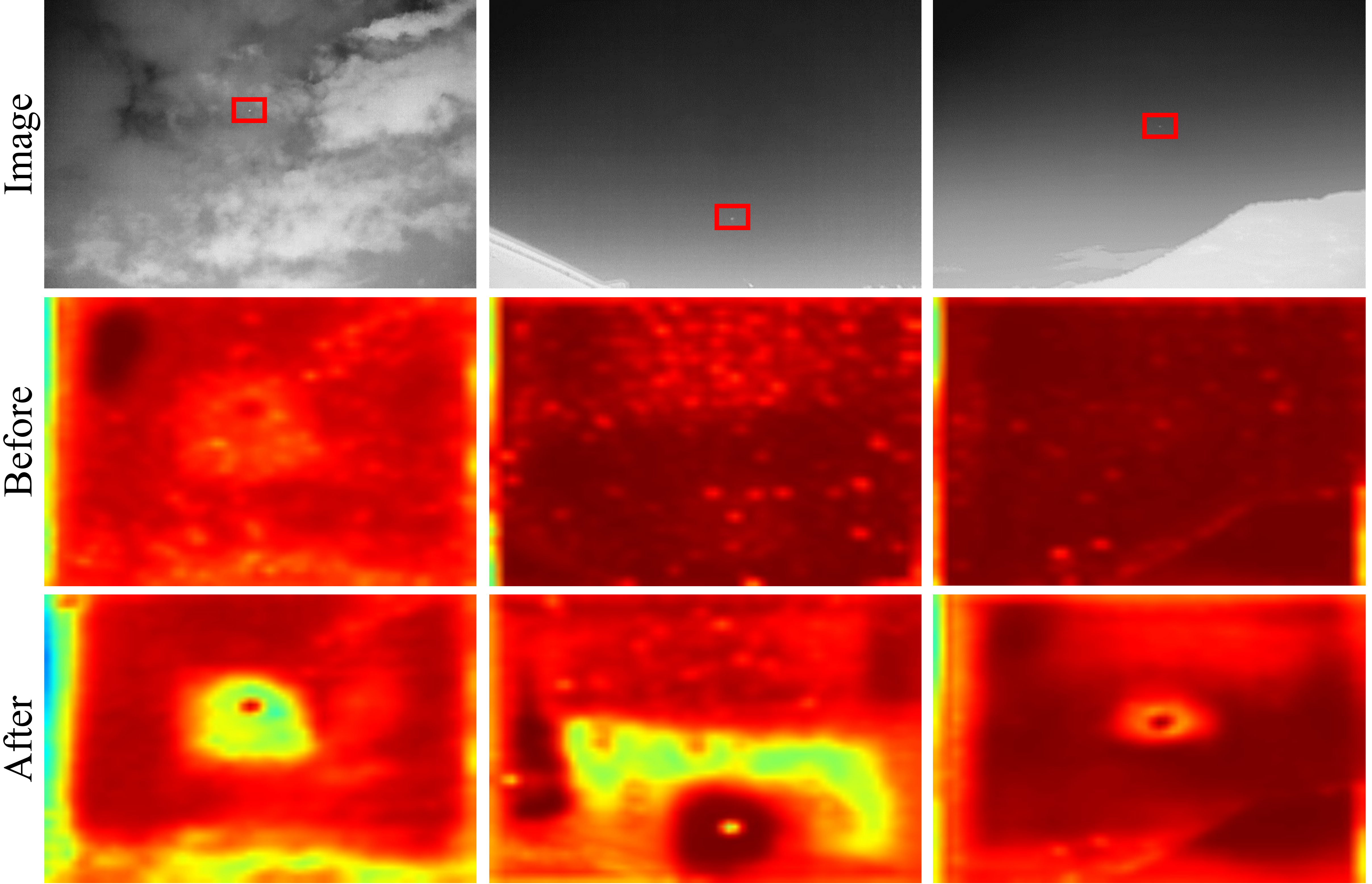}
	\caption{Visualization of feature maps before and after coherent-motion-guided local anomaly reasoning module using PCA-based channel reduction. The proposed framework effectively suppresses coherent-motion-induced false responses while enhancing target-sensitive local motion regions.}
	\label{fig:fusion}
\end{figure}
As illustrated in Fig.~\ref{fig:fusion}, the proposed reasoning strategy effectively suppresses coherent-motion-consistent false responses while preserving target-sensitive localized anomalies. Consequently, the implicit branch can better focus on residual localized motion anomalies associated with infrared small targets. Overall, the proposed framework demonstrates that coherent-motion priors are essential not only for suppressing false-alarm responses, but also for stabilizing local motion reasoning under severe motion entanglement. Unlike previous model-driven approaches \cite{wang2026improved} that adopt a sequential ``alignment-then-detection'' pipeline to handle dynamic scenes, these methods are inherently cascaded and thus suffer from accumulated errors propagated across stages. In contrast, our coherent-motion-guided local anomaly reasoning module enables a \textit{\textbf{closed-loop}} interaction between the two branches, allowing mutual correction and refinement of motion representations. This design effectively mitigates error accumulation and further improves robustness in non-rigid scenes where accurate alignment is difficult to achieve, leading to more stable and reliable motion reasoning under complex dynamic conditions.

\subsection{Loss Function}
Besides the self-supervised correspondence objective, the proposed framework employs the YOLOX detection head and is trained with the standard detection loss:
\begin{equation}
	\mathcal{L}_{det}
	=
	\lambda_{reg}\mathcal{L}_{reg}
	+
	\lambda_{obj}\mathcal{L}_{obj}
	+
	\lambda_{cls}\mathcal{L}_{cls},
\end{equation}
where $\mathcal{L}_{reg}$, $\mathcal{L}_{obj}$, and $\mathcal{L}_{cls}$ denote the regression, objectness, and classification losses, respectively, and the loss weights follow the default YOLOX setting \cite{yolox2021}.

\section{Experiment}\label{Section:Experiment}
\subsection{Experimental Setup}
\subsubsection{Datasets} To validate the proposed method, we conduct experiments on two widely used infrared small target detection benchmarks: IRDST and ITSDT-15K. IRDST represents a long-range air-to-ground scenario, where UAV targets appear as tiny point-like objects under dynamic backgrounds and significant relative motion. In contrast, ITSDT-15K focuses on a ground-to-air scenario with moving vehicle targets, relatively stable backgrounds, and partial occlusion.

Following the official protocols \cite{RDIAN,duan2024triple}, IRDST is split into 42 training videos (20,398 frames) and 43 testing videos (20,258 frames), while ITSDT-15K contains 10,000 training images from 40 videos and 5,000 testing images from 20 videos. Training and test sets are sampled from different sequences with diverse backgrounds to ensure fair evaluation and robust generalization assessment.

\subsubsection{Evaluation Metrics} For performance evaluation, we adopt several widely used metrics in object detection, including Parameters, FPS (input size $512\times512$, batchsize 1), Precision (Pr), Recall (Re), F1-score, $\text{mAP}_{50}$, as well as Precision–Recall (PR) curves, to comprehensively assess the effectiveness of the proposed model.

\subsubsection{Implementation Details} In implementation, we followed the same settings as the baseline. The input image resolution was fixed at $512\times512$. We trained our model for 100 epochs with a batch size of 4. The initial learning rate was set to 0.01, and stochastic gradient descent (SGD) was adopted as the optimizer, with a momentum of 0.937, a weight decay of $5\times10^{-4}$, and a learning rate decay factor of 0.1. During testing, only the predicted bounding boxes with confidence scores greater than 0.001 were retained. The intersection-over-union (IoU) threshold for non-maximum suppression (NMS) was set to 0.65. All experiments were conducted on a single NVIDIA 3080TI GPU.

\subsection{Comparison with State-of-the-Arts}
\begin{table*}[t!]
	\caption{Detection results achieved by different SOTA methods. The best results are in \textbf{bold}, and the second-best results are \underline{underlined}.}\label{Tab:SOTA}
	\centering
	\setlength{\tabcolsep}{6.0pt}
	\definecolor{mintgreen}{RGB}{204, 255, 204}   
	\definecolor{skyblue}{RGB}{204, 238, 255}     
	\definecolor{peachpink}{RGB}{255, 221, 238}   
	\resizebox{\textwidth}{!}{\begin{tabular}{@{}ccl|c|cccc|cccc|rr}
			\noalign{\hrule height 1pt}
			\multicolumn{3}{c|}{\multirow{2}*{Methods}} & \multicolumn{1}{c|}{\multirow{2}*{T}} & \multicolumn{4}{c}{\textbf{ITSDT-15K} \colorbox{skyblue}{[Static Scenes]}} & \multicolumn{4}{c|}{\textbf{IRDST} \colorbox{peachpink}{[Dynamic Scenes]}} & \multirow{2}*{Params}  & \multirow{2}*{FPS} \\
			\cline{5-12}
			\multicolumn{3}{c|}{~} & \multicolumn{1}{c|}{~}  & $\textbf{mAP}_\textbf{50}$ & \textbf{Pr} & \textbf{Re}& \textbf{F1} &$\textbf{mAP}_\textbf{50}$  & \textbf{Pr} & \textbf{Re}  & \textbf{F1}                  \\ \noalign{\hrule height 1pt}
			& \multicolumn{1}{c}{\multirow{18}{*}{\rotatebox{90}{\textit{Single-Frame}}}}  & ACM \cite{Dai_2021_WACV} \textit{(WACV'21)} & N & 55.38 & 78.37 & 71.69 & 74.88 & 52.40 & 76.33 & 69.32 & 72.66 & 3.04M & 22.57 \\
			& \multicolumn{1}{c}{}  & YOLOX \cite{yolox2021} \textit{(ARXIV'21)} & N & 71.95 & 83.43 & 87.35 & 85.34 & 63.15 & 84.56 & 75.35 & 79.69 & 7.259M  & 111\\
			& \multicolumn{1}{c}{}  & ISNet \cite{ISNet} \textit{(CVPR'22)}  & N & 62.29 & 83.46 & 75.32 & 79.18 & 59.78 & 80.24 & 75.08 & 77.58 & 3.48M  & 8.87 \\
			& \multicolumn{1}{c}{}  & UIUNet \cite{UIUNet} \textit{(TIP'22)} & N & 65.15 & 84.07 & 78.39 & 81.13 & 56.38 & 80.95 & 70.29 & 75.25 & 53.06M  & 3.63 \\
			& \multicolumn{1}{c}{}  & AGPCNet \cite{AGPCNet} \textit{(TAES'23)} & N & 67.27 & 91.19 & 74.77 & 82.16 & 59.21 & 79.47 & 75.51 & 77.44 & 14.88M  & 4.79 \\
			& \multicolumn{1}{c}{}  & RDIAN \cite{RDIAN} \textit{(TGRS'23)} & N & 68.49 & 90.56 & 76.06 & 82.68 & 59.08 & 77.99 & 76.35 & 77.16 & 2.74M  & 20.52 \\
			& \multicolumn{1}{c}{}  & DNANet \cite{DNANet} \textit{(TIP'23)} & N & 70.46 & 88.55 & 80.73 & 84.46 & 63.61 & 82.92 & 77.48 & 80.11 & 7.22M  & 4.78\\
			& \multicolumn{1}{c}{}  & SIRST5K \cite{lu2024sirst} \textit{(TGRS'24)} & N & 61.52 & 86.95 & 71.32 & 78.36 & 52.28 & 76.12 & 69.07 & 72.42 & 11.48M  & 7.37 \\
			& \multicolumn{1}{c}{}  & RPCANet \cite{RPCANet} \textit{(WACV'24)} & N & 62.28 & 81.46 & 77.10 & 79.22 & 56.50 & 77.77 & 73.80 & 75.73 & 3.21M  & 15.85  \\
			& \multicolumn{1}{c}{}  & YOLOH \cite{10471352}  \textit{(TIP'24)} & N & 54.99 & 79.98 & 69.67 & 74.47 & 61.57 & 80.48 & 77.76 & 79.10 & 7.542M  & 98.23  \\
			& \multicolumn{1}{c}{}  & RT-DETR \cite{10657220}  \textit{(CVPR'24)} & N & 57.17 & 77.15 & 55.01 & 64.22 & 68.95 & 70.01 & 68.79 & 69.39 & 42.8M  & 102.8  \\
			& \multicolumn{1}{c}{}  & MSHNet \cite{MSHNet} \textit{(CVPR'24)}& N  & 60.82 & 89.69 & 68.44 & 77.64 & 63.21 & 82.31 & 77.64 & 79.91 & 6.59M  & 18.55 \\
			& \multicolumn{1}{c}{}  & MLPNet \cite{10793117} \textit{(TGRS'25)}& N  & 53.76 & 74.06 & 73.19 & 73.63 & 57.48 & 80.36 & 72.14 & 76.03 & 10.79M  & 5.93  \\
			& \multicolumn{1}{c}{}  & L2SKNet \cite{10813615}  \textit{(TGRS'25)} & N & 66.07 & 90.75 & 73.72 & 81.35 & 61.19 & 77.85 & 79.98 & 78.90 & 3.42M  & 40.63 \\
			& \multicolumn{1}{c}{}  & PConv \cite{yang2025pinwheel} \textit{(AAAI'25)} & N & 76.02 & 89.96 & 85.18 & 87.50 & 66.43 & 88.93 & 75.54 & 81.69 & 6.59M  & 10.01
			\\
			& \multicolumn{1}{c}{}  & Hyper-YOLO \cite{feng2024hyper} \textit{(TPAMI'25)} & N & 77.12 & 89.47 & 61.15 & 72.64 & 69.90 & 86.67 & 81.59 & 84.05 & 14.8M  & 127 \\
			& \multicolumn{1}{c}{}  & HR-SemNet \cite{11361356} \textit{(TIP'26)}& N  & 74.47 & 73.00 & 77.50 & 75.20 & 71.55 & 73.71 & 70.65 & 72.17 & 8.7M  & 117 \\
			& \multicolumn{1}{c}{}  & NS-FPN \cite{yuan2025ns}  \textit{(CVPR'26)}& N  & 77.12 & 88.76 & 86.31 & 88.18 & 67.80 & \textbf{89.27} & 76.68 & 82.51 & 7.89M  & 11.17
			\\
			\cline{2-14} 
			& \multicolumn{1}{c}{\multirow{22}{*}{\rotatebox{90}{\textit{Multi-Frame}}}}
			& TROI  \cite{gong2021temporal} \textit{(AAAI'21)} & I & 68.14 & 91.00 & 75.36 & 82.45 & 64.67 & 87.46 & 74.90 & 80.70 & 97.20M  & 13.6 \\
			& \multicolumn{1}{c}{} & ST-Adapter \cite{NEURIPS2022_a92e9165} \textit{(NeurIPS'22)} & I  & 78.10 &\textbf{93.40} & 84.73 & 88.85 & 72.22 & 87.52 & 83.50 & 85.46 & 12.73M  & 9.76 \\
			& \multicolumn{1}{c}{}  &  TransVOD \cite{9960850} \textit{(TPAMI'23)} & I & 63.59 & 88.37 & 70.34 & 78.30 & 58.68 & 81.25 & 69.09 & 74.68 & 39.83M  & 13.82 \\
			& \multicolumn{1}{c}{}  & TMP \cite{zhu2024tmp} \textit{(ESWA'24)}  & I& 77.50 & 90.65 & 86.89 & 88.73 & 70.03 & 86.70 & 81.41 & 83.97 & 16.41M  & 6.91 \\ 
			& \multicolumn{1}{c}{}  & ST-Trans \cite{10409231} \textit{(TGRS'24)}  & I & 76.02 & 89.96 & 85.18 & 87.50 & 70.04 & 88.21 & 80.01 & 83.91 & 38.13M  & 3.9 \\
			& \multicolumn{1}{c}{}  & Tridos \cite{duan2024triple} \textit{(TGRS'24)}  & I & 76.72 & 91.81 & 84.63 & 88.07 & 73.72 & 84.49 & 89.35 & 86.85 & 14.13M  & 13.71 \\
			& \multicolumn{1}{c}{}  & SSTNet \cite{chen2024sstnet} \textit{(TGRS'24)}  & I & 76.96 & 91.05 & 85.29 & 88.07 & 71.55 & 88.56 & 81.92 & 85.11 & 11.95M  & 12.35 \\
			& \multicolumn{1}{c}{}  & STMENet \cite{peng2025moving} \textit{(EAAI'25)}  & I & 77.33 & 92.42 & 84.35 & 88.21 & 73.40 & 87.78 & 84.22 & 85.96 & 9.85M  & 11.95 \\
			& \multicolumn{1}{c}{}  & S2MVP \cite{10824834}  \textit{(TGRS'25)}  & I & 78.17 & 88.49 & 88.88 & 88.69 & 72.15 & 85.36 & 86.87 & 86.11 & 52.74M  & 28.76 \\
			& \multicolumn{1}{c}{}  & MFENet \cite{11145128}  \textit{(TGRS'25)}  & E & 50.70 & 82.11 & 62.70 & 71.10 & 65.31 & 80.19 & 82.07 & 81.12 & 27.58M  & 4.74 \\
			& \multicolumn{1}{c}{}& DTUM \cite{li2023direction} \textit{(TNNLS'25)}  & E & 67.97 & 77.95 & 88.28 & 82.79 & 71.48 & 82.87 & 87.79 & 85.26 & 9.64M  & 14.24 \\
			& \multicolumn{1}{c}{}& EMIP \cite{10989627} \textit{(TIP'25)}  & E & 73.50 & 91.22 & 81.05 & 85.83 & 71.11 & 86.98 & 82.25 & 84.55 & 46.03M  & 20.85\\
			& \multicolumn{1}{c}{}  & DST-Adapter\cite{pei2025d2st} \textit{(ICCV'25)} & I & 76.80 & 90.87 & 85.36 & 88.03 & 70.34 & 86.15 & 82.44 & 84.25 & 12.38M  & 9.33 \\
			& \multicolumn{1}{c}{}  & MoPKL \cite{chen2025motion} \textit{(AAAI'25)}  & M & 79.78 & \underline{93.29} & 86.80 & 89.92 & 74.54 & \underline{89.04} & 84.74 & 86.84 & 9.46M  & 10.03 \\
			& \multicolumn{1}{c}{}  &  BeltCrack \cite{HUANG2026113598} \textit{(PR'26)}  & I & 74.90 & 89.67 & 84.82 & 87.18 & 72.91 & 86.64 & 85.12 & 85.82 & 20.70M  & 16.70 \\
			& \multicolumn{1}{c}{}  &  ADSUNet \cite{zhang2026adsunet} \textit{(PR'26)}  & E & 46.55 & 70.79 & 66.17 & 68.40 & \underline{75.00} & 86.76 & 87.41 & 87.08 & 26.06M  & 20.82 \\
			& \multicolumn{1}{c}{}  &  MISTNet \cite{gao2026mist}  \textit{(TIP'26)}  & I & 57.20 & 79.84 & 72.50 & 75.99 & 64.32 & 80.87 & 80.19 & 80.53 & 3.914M  &  6.73\\
			& \multicolumn{1}{c}{}  &  CoMoE \cite{duan2026cross}   \textit{(AAAI'26)}  & I & \underline{80.40} & 88.15 & \underline{92.98} & \underline{90.50} & 63.51 & 73.65 & 87.44 & 79.95 & 19.61M  & 12.73 \\
			& \multicolumn{1}{c}{}  & SeVIL \cite{duan2026sevil} \textit{(AAAI'26)}  & M & 80.18 & 88.96 & 91.46 & 90.20 & 74.62 & 86.53 & 87.82 & \underline{87.17} & 40.48M  & 21.7 \\
			& \multicolumn{1}{c}{}  &  CHAL \cite{Duan_2026_CVPR} \textit{(CVPR'26)}  & E & 74.40 & 89.38 & 84.35 & 86.79 & 71.55 & 81.07 & \underline{89.61} & 85.12 & 15.69M  & 12.96 \\
			& \multicolumn{1}{c}{}  &  DMRNet \textit{(Ours)}  & D & \textbf{82.59} & 89.58 & \textbf{93.98} & \textbf{91.72} & \textbf{79.13} & 87.53 & \textbf{91.23} & \textbf{89.34} & 13.75M  & 19.74 \\
			\noalign{\hrule height 1pt}
			\multicolumn{14}{l}{\footnotesize{
					T: motion modeling type (N: no motion, I: implicit, E: explicit, M: vision-language multimodal, D: decoupled motion modeling).
			}}
	\end{tabular}}
\end{table*}

\subsubsection{Quantitative Evaluation}
The quantitative results in Tab. \ref{Tab:SOTA} reveal three notable observations. First, multi-frame methods consistently outperform single-frame approaches, confirming the effectiveness of temporal information. Second, implicit motion modeling generally yields better performance than explicit motion modeling, suggesting that the motion characteristics of infrared small targets in dynamic scenes are difficult to describe with handcrafted motion formulations. Third, vision-language multimodal methods further improve detection accuracy by introducing semantic priors for motion understanding. Nevertheless, these gains are often accompanied by substantially increased computational cost. To address this challenge, our method disentangles complex motion into globally coherent and locally independent components, and performs motion reasoning to suppress false alarms while enhancing target responses, thereby achieving state-of-the-art performance with superior efficiency.

In dynamic scenes, the proposed method achieves 79.13\% mAP$_{50}$ with only 13.75M parameters and a real-time inference speed of 19.74 FPS. In contrast, among existing implicit motion modeling approaches, SeVIL requires external language guidance describing target motion and a substantially larger model size of 40.48M parameters to reach 74.62\% mAP$_{50}$. Among purely visual methods, Tridos achieves competitive performance of 73.72\% mAP$_{50}$ by introducing implicit spatiotemporal–frequency domain modeling; however, this design also leads to a reduced inference speed of 13.71 FPS and a parameter count of 14.13M. These results clearly highlight the superiority of our motion decoupling framework in achieving a better balance between detection accuracy, computational efficiency, and model compactness.

Even in static scenes, the proposed method still achieves 82.59\% mAP$_{50}$. This is mainly attributed to the fact that infrared scenes contain numerous false-alarm sources. By leveraging motion decoupling and conditional motion reasoning, the cooperative interaction between the two branches effectively suppresses false alarms while enhancing true target responses.
In contrast, existing methods achieve only 80.18\% mAP$_{50}$ even when equipped with external language guidance describing target motion. This limitation arises because target motion is often difficult to accurately describe in natural language, making precise characterization challenging. Moreover, language descriptions typically focus on the target itself while lacking explicit modeling of background-induced false-alarm sources, which further restricts their ability to disambiguate complex infrared scenes.
\begin{figure}
	\centering
	\includegraphics[width=\linewidth]{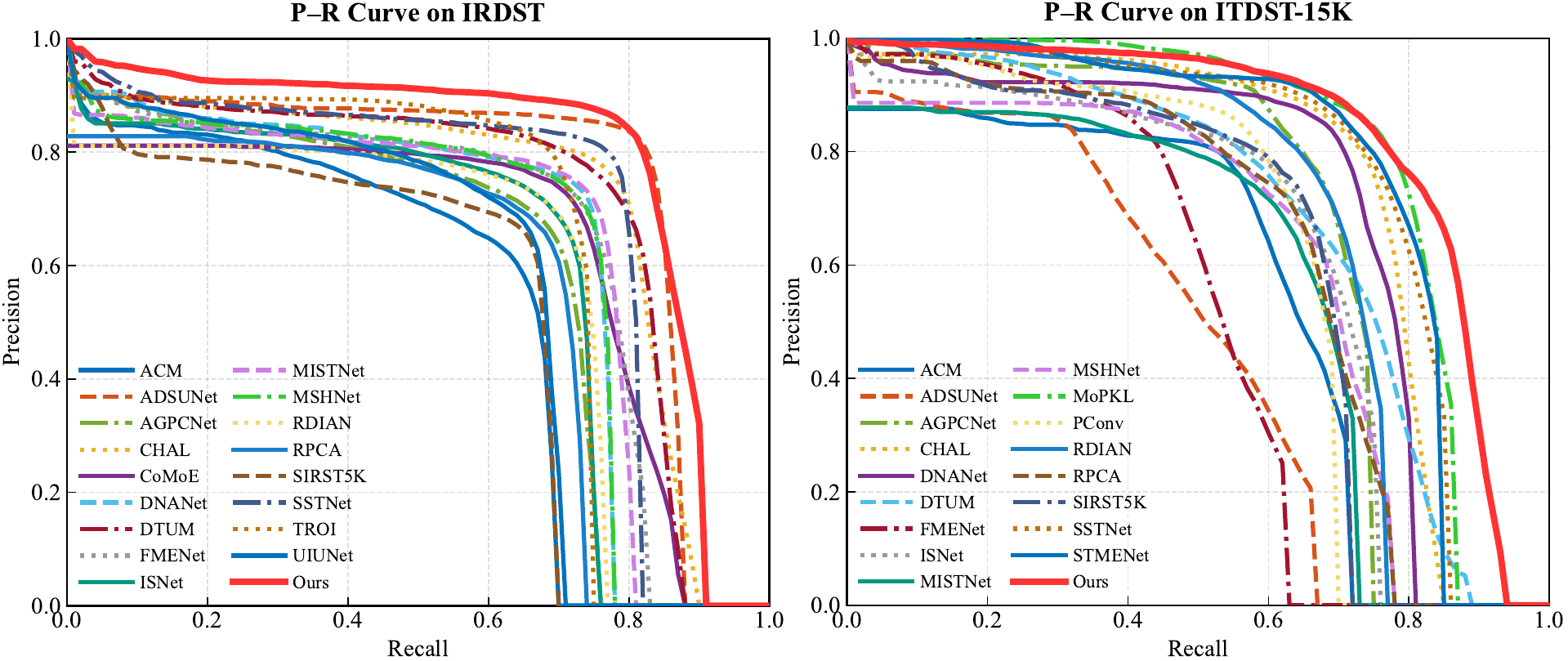}
	\caption{Precision–Recall Curves on IRDST dataset and ITDST-15K dataset.}
	\label{fig:PR}
\end{figure}
The P–R curve in Fig. \ref{fig:PR} further demonstrates the excellent precision‑recall balance of our approach.
\begin{table*}[!t]
	\renewcommand\arraystretch{1.}
	\centering
	\caption{Performance Comparison under Different Average Target Velocities on IRDST Dataset.}
	\label{Tab:VE}
	\resizebox{\textwidth}{!}{\begin{tabular}{c|cc|cc|cc|cc|cc|cc}
			\noalign{\hrule height 1pt}
			 \multirow{2}{*}{Avg Target Motion Intensity}              & \multicolumn{2}{c|}{Ours}      & \multicolumn{2}{c|}{MISTNet \cite{gao2026mist}} & \multicolumn{2}{c|}{CHAL \cite{Duan_2026_CVPR}} & \multicolumn{2}{c|}{ADSUNet \cite{zhang2026adsunet}} & \multicolumn{2}{c|}{EMIP \cite{10989627}} & \multicolumn{2}{c}{DST-Adapter \cite{pei2025d2st}} \\ \cline{2-13}
			\multicolumn{1}{c|}{} & $\textbf{mAP}_\textbf{50}$    & \textbf{F1}  & $\textbf{mAP}_\textbf{50}$    & \textbf{F1} & $\textbf{mAP}_\textbf{50}$    & \textbf{F1}     & $\textbf{mAP}_\textbf{50}$            & \textbf{F1}   & $\textbf{mAP}_\textbf{50}$            & \textbf{F1}    & $\textbf{mAP}_\textbf{50}$            & \textbf{F1}                                      \\ \noalign{\hrule height 1pt}
		 4.660 pixels/frame & \textbf{99.05} & \textbf{99.91} & 75.10 & 87.05 & 89.21 & 94.91 & 87.70 & 94.08  & 96.89 & 98.87 & \underline{99.00} & \underline{99.90} \\
			 12.40 pixels/frame & \textbf{87.75} & \textbf{93.85}  & 69.21 & 83.85  & 76.20 & 87.73 &\underline{82.40} & \underline{91.20}& 71.53 & 84.82 & 68.17 & 81.73  \\
			\cline{1-12} \rowcolor[rgb]{0.9,0.9,0.9}
			33.12 pixels/frame & \textbf{72.30} & \textbf{85.23}  & \underline{12.84} & \underline{34.47} & 0.0 & 0.0 & 9.1 & 29.02 & 0.0 & 0.0 & 1.8 & 10.2 \\
			 \noalign{\hrule height 1pt}
	\end{tabular}}
\end{table*}
A more fine-grained comparison is presented in Tab. \ref{Tab:VE}. When the average target velocity reaches 33.12 pixels/frame, the performance of MISTNet drops dramatically to 12.84\% mAP$_{50}$, while CHAL, ADSUNet, EMIP and DST-Adapter further decrease to 0.0\% mAP$_{50}$, 9.1\% mAP$_{50}$, 0.0\% mAP$_{50}$ and 1.8\% mAP$_{50}$, respectively. In contrast, our method still maintains 72.30\% mAP$_{50}$, demonstrating its superior capability in handling fast motion patterns.
\subsubsection{Qualitative Evaluation}
\begin{figure*}[!t]
	\centering
	\includegraphics[width=\linewidth]{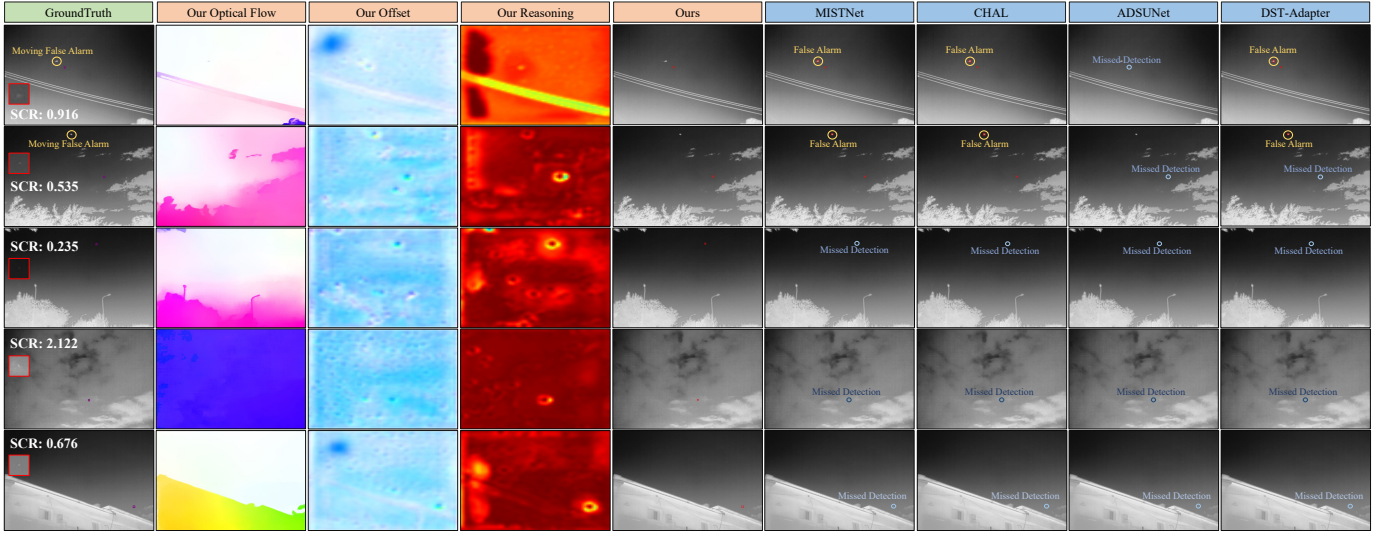}
	\caption{Qualitative visualization results. In dynamic scenes, the proposed optical flow branch effectively captures coherent global motion, while the deformable offsets focus on local target motion. Through the proposed motion reasoning strategy, the real target motion can be accurately identified, and the final motion reasoning results effectively suppress background-induced false alarms. For better visualization, the feature maps refined by the proposed motion reasoning module are projected into low-dimensional representations using PCA. In contrast, methods relying solely on implicit motion modeling tend to either miss true targets or mistakenly regard motion-induced false alarms as targets. [For better visual presentation, please zoom in the images.]}
	\label{fig:VisualResult}
\end{figure*}
The qualitative results are illustrated in Fig. \ref{fig:VisualResult}, which presents visualization comparisons of infrared small target detection under challenging scenarios. The results demonstrate the effectiveness of our method in handling complex background clutter interference and complicated target motion patterns in dynamic scenes. As shown in the first two rows, it suppresses motion-induced false alarms caused by imaging platform dynamics while preserving sensitivity to weak targets, whereas existing methods either miss true targets or misclassify motion-related background responses.

\subsection{Ablation Study}

\subsubsection{Ablation Study on Module-wise Performance Gain}
\begin{table}[!t]
	\renewcommand\arraystretch{1.}
	\centering
	\caption{Ablation Study on Module-wise Performance Gain on IRDST Dataset. ``E'' denotes the explicit motion branch, ``I'' denotes the implicit motion branch, and ``R'' denotes the coherent-motion-guided local anomaly reasoning module.}
	\label{Tab:AB_all}
	\resizebox{0.48\textwidth}{!}{\begin{tabular}{c|c|c|cccc}
			\noalign{\hrule height 1pt}
			\textbf{E}      &\textbf{I} &\textbf{R}              & $\textbf{mAP}_\textbf{50}(\%)$$\uparrow$    & $\textbf{Pr}(\%)$$\uparrow$    & $\textbf{Re}(\%)$$\uparrow$            & \textbf{F1}(\%)$\uparrow$                                      \\ \noalign{\hrule height 1pt}
			\XSolidBrush & \XSolidBrush & \XSolidBrush & 63.15 & 84.56 & 75.35 & 76.69 \\
			\cline{1-7} 
			\CheckmarkBold & \XSolidBrush & \XSolidBrush & 73.00 & 88.36 & 83.32 & 85.77 \\
			\cline{1-7} 
			\XSolidBrush & \CheckmarkBold & \XSolidBrush & 69.68 & 87.27 & 80.53 & 83.76 \\
			\cline{1-7} 
			\CheckmarkBold & \XSolidBrush & \CheckmarkBold & 74.70 & \textbf{89.21} & 84.40 & 86.74 \\
		\cline{1-7}
		\CheckmarkBold	& \CheckmarkBold & \CheckmarkBold & \textbf{79.13} & 87.53 & \textbf{91.23} & \textbf{89.34}  \\ \noalign{\hrule height 1pt}
	\end{tabular}}
\end{table}
As reported in Tab. \ref{Tab:AB_all}, the baseline achieves only 63.15\% mAP$_{50}$. Introducing only the implicit motion branch improves the performance to 69.68\% mAP$_{50}$, indicating that local motion modeling alone is still insufficient for handling highly coupled motion patterns in complex infrared scenes. Using only the explicit motion branch further improves the performance to 73.00\% mAP$_{50}$, since globally coherent motion priors help suppress substantial background interference. However, optical flow estimation is inherently less effective at capturing the subtle motions of infrared small targets, leading to a clear performance bottleneck. By further introducing the coherent-motion-guided local anomaly reasoning module, the performance increases to 74.70\% mAP$_{50}$, demonstrating that coherent-motion-guided refinement effectively suppresses coherent-motion-induced false responses within the detection backbone. Finally, integrating the explicit motion branch, implicit motion branch, and conditional motion reasoning module together achieves the best performance of 79.13\% mAP$_{50}$. This validates that the proposed framework successfully enables motion disentanglement and collaborative motion reasoning, thereby effectively adapting to complex coupled motion patterns.
\subsubsection{Ablation study of different components in the explicit motion branch}
\begin{table}[!t]
	\renewcommand\arraystretch{1.}
	\centering
	\caption{Ablation study of different components in the explicit motion branch on IRDST Dataset. ``S'' denotes the self-supervised learning strategy, ``B'' denotes the learnable band-pass filter, and ``P'' denotes phase enhancement.}
	\label{Tab:AB_E}
	\resizebox{0.48\textwidth}{!}{\begin{tabular}{c|c|c|cccc}
			\noalign{\hrule height 1pt}
			\textbf{S}      &\textbf{B}  &\textbf{P}              & $\textbf{mAP}_\textbf{50}(\%)$$\uparrow$    & $\textbf{Pr}(\%)$$\uparrow$    & $\textbf{Re}(\%)$$\uparrow$            & \textbf{F1}(\%)$\uparrow$                                      \\ \noalign{\hrule height 1pt}
			\XSolidBrush & \XSolidBrush & \XSolidBrush & 68.13 & 86.40 & 79.42 & 82.76 \\
			\cline{1-7} 
			\CheckmarkBold & \XSolidBrush & \XSolidBrush & 74.98 & 88.36 & 85.63 & 86.98 \\
			\cline{1-7} 
			\CheckmarkBold & \CheckmarkBold & \XSolidBrush & 77.20 & \textbf{90.10} & 86.69 & 88.36 \\
			\cline{1-7} 
			\CheckmarkBold & \XSolidBrush & \CheckmarkBold & 77.23 & 89.26 & 87.27 & 88.25 \\
			\cline{1-7}
			\CheckmarkBold	& \CheckmarkBold & \CheckmarkBold & \textbf{79.13} & 87.53 & \textbf{91.23} & \textbf{89.34}  \\ \noalign{\hrule height 1pt}
	\end{tabular}}
\end{table}
The qualitative and quantitative results of each component in the explicit motion branch are presented in Tab. \ref{Tab:AB_E} and Fig. \ref{fig:op}. With only the multi-scale prompt, the performance reaches 68.13\% mAP$_{50}$. After introducing the self-supervised learning strategy, the performance is significantly improved to 74.98\% mAP$_{50}$, highlighting that self-supervised signals during training can effectively preserve motion representation awareness.
On this basis, incorporating the band-pass filter further boosts the performance to 77.20\% mAP$_{50}$, as it effectively suppresses strong high-frequency noise and low-frequency irrelevant information in infrared images, thereby enhancing representation quality. Introducing phase enhancement alone yields a comparable improvement to 77.23\% mAP$_{50}$, since phase information emphasizes structural cues that are beneficial for capturing motion characteristics. However, since useful information is not distributed uniformly across all frequency bands, combining phase enhancement with the band-pass filter ultimately achieves the best performance of 79.13\% mAP$_{50}$.
In addition, Tab. \ref{Tab:PEFT} and comparisons with different parameter-efficient fine-tuning methods further validate the effectiveness and rationality of the proposed explicit motion branch design.

\subsubsection{Ablation study of different components in the implicit motion branch}
The qualitative and quantitative results demonstrating the effectiveness of each component in the implicit motion branch are presented in Tab. \ref{Tab:Offset} and Fig. \ref{fig:offset}. When the knowledge guidance from the explicit motion branch is removed, the performance drops from 79.13\% mAP$_{50}$ to 77.80\% mAP$_{50}$.
In addition, the proposed method exhibits more stable performance compared to existing offset estimation-based approaches. This is because prior methods directly supervise deformable offset estimation using optical flow results. However, optical flow estimation is inherently limited in capturing small target motions, which leads to unreliable supervision signals and consequently hinders the implicit motion branch from effectively modeling localized motion patterns associated with potential targets.
\subsubsection{Ablation study of different components in the coherent-motion-guided local anomaly reasoning module}
\begin{table}[!t]
	\renewcommand\arraystretch{1.}
	\centering
	\caption{Ablation study of different components in the coherent-motion-guided local anomaly reasoning module on IRDST Dataset. ``S'' denotes the subtraction strategy, ``M'' denotes the multi-scale large-kernel attention mechanism, and ``Q'' denotes the self-attention-based refinement of the implicit motion branch.}
	\label{Tab:AB_R}
	\resizebox{0.48\textwidth}{!}{\begin{tabular}{c|c|c|cccc}
			\noalign{\hrule height 1pt}
			\textbf{S}      &\textbf{M}  &\textbf{Q}              & $\textbf{mAP}_\textbf{50}(\%)$$\uparrow$    & $\textbf{Pr}(\%)$$\uparrow$    & $\textbf{Re}(\%)$$\uparrow$            & \textbf{F1}(\%)$\uparrow$                                      \\ \noalign{\hrule height 1pt}
			\CheckmarkBold & \XSolidBrush & \XSolidBrush & 70.71 & 87.63 & 81.57 & 84.49 \\
			\cline{1-7} 
			\XSolidBrush & \CheckmarkBold & \XSolidBrush & 76.57 & \textbf{90.27} & 85.94 & 88.05 \\
			\cline{1-7} 
			\XSolidBrush & \XSolidBrush & \CheckmarkBold & \textbf{79.13} & 87.53 & \textbf{91.23} & \textbf{89.34}  \\ \noalign{\hrule height 1pt}
	\end{tabular}}
\end{table}
To verify the effectiveness of the coherent-motion-guided local anomaly reasoning module, we first examine whether global and local motions can be linearly decoupled. To this end, we replace the cross-attention mechanism with a simple subtraction operation. As shown in Tab. \ref{Tab:AB_R}, the performance drops significantly from 79.13\% mAP$_{50}$ to 70.71\% mAP$_{50}$, even falling below that of the explicit motion branch alone.
Furthermore, we replace the self-attention-based refinement of the implicit motion branch with an large selective kernel attention mechanism \cite{Li_2024_IJCV}. In this case, the performance further decreases to 76.57\% mAP$_{50}$. This is because the refinement module is designed to aggregate false-alarm features that are consistent with globally coherent motion patterns. Although large-kernel attention can enlarge the effective receptive field, it primarily emphasizes local inductive bias rather than global feature correlation, thereby limiting its ability to model globally consistent motion dependencies.
\subsubsection{Background Dynamic Variation Analysis}
\begin{figure}[!t]
	\centering
	\includegraphics[width=\linewidth]{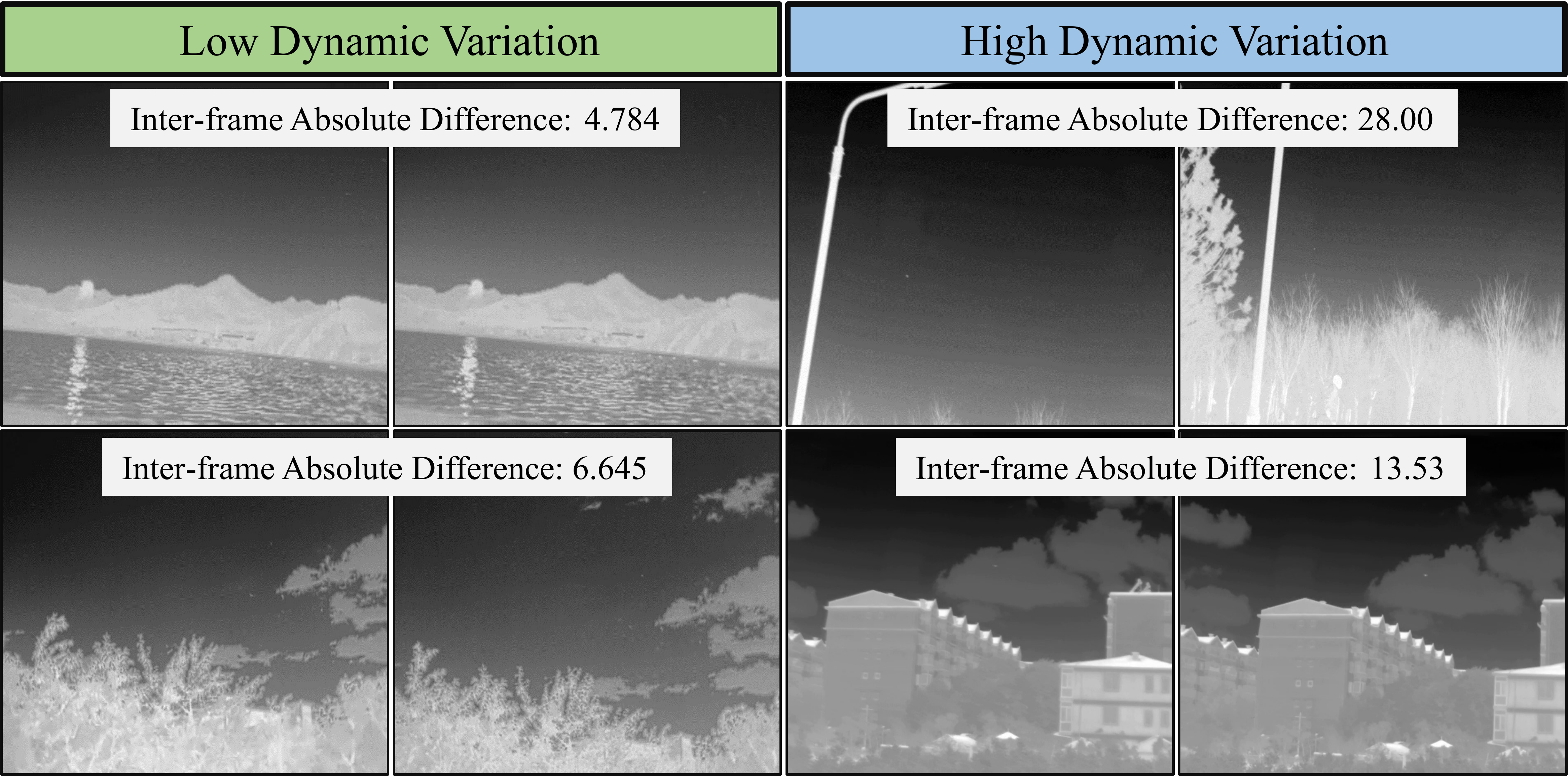}
	\caption{Illustration of Background Dynamic Variation. Representative sequences with low and high Dynamic Variation are presented to highlight the substantial difference between relatively stable and highly dynamic scenarios. Notably, even under low Dynamic Variation conditions, non-rigid background motions (\eg, vegetation) may still induce noticeable false alarms, posing considerable challenges for reliable infrared small target detection. This partition is further used to evaluate the robustness of different motion modeling strategies under varying dynamic complexities.}
	\label{fig:DynamicVariation}
\end{figure}

To evaluate robustness under different background dynamics, sequences are partitioned into low-, high-, and extreme-dynamic subsets according to the average inter-frame absolute difference (AIFAD), as illustrated in Fig.~\ref{fig:DynamicVariation}. Representative examples show that even relatively stable scenes may contain non-rigid motions (\eg, vegetation), which can still produce noticeable false alarms and challenge reliable infrared small target detection.
\begin{table*}[!t]
	\renewcommand\arraystretch{1.}
	\centering
	\caption{Performance Comparison under Different Background Dynamic Variation on IRDST Dataset.}
	\label{Tab:DVTr}
	\resizebox{\textwidth}{!}{\begin{tabular}{c|c|cc|cc|cc|cc|cc|cc}
			\noalign{\hrule height 1pt}
			\multirow{2}{*}{Background Dynamic Variation}  & \multirow{2}{*}{Type}            & \multicolumn{2}{c|}{Ours}      & \multicolumn{2}{c|}{MISTNet \cite{gao2026mist}} & \multicolumn{2}{c|}{CHAL \cite{Duan_2026_CVPR}} & \multicolumn{2}{c|}{ADSUNet \cite{zhang2026adsunet}} & \multicolumn{2}{c|}{EMIP \cite{10989627}} & \multicolumn{2}{c}{CoMoE \cite{duan2026cross}} \\ \cline{3-14}
			\multicolumn{1}{c|}{} & \multicolumn{1}{c|}{} & $\textbf{mAP}_\textbf{50}$    & \textbf{F1}  & $\textbf{mAP}_\textbf{50}$    & \textbf{F1} & $\textbf{mAP}_\textbf{50}$    & \textbf{F1}     & $\textbf{mAP}_\textbf{50}$            & \textbf{F1}   & $\textbf{mAP}_\textbf{50}$            & \textbf{F1}    & $\textbf{mAP}_\textbf{50}$            & \textbf{F1}                                      \\ \noalign{\hrule height 1pt}
			 AIFAD = 3.5342 & Low-Dynamic & \textbf{100.0} & \textbf{100.0} & 91.80 & 95.89 & 90.65 & 95.34 & \underline{94.21} & \underline{97.14} &  94.10 & 97.34 & 92.85  &  96.76  \\
			 AIFAD = 16.018 & High-Dynamic & \textbf{96.20} & \textbf{98.56} & 89.90 &  94.84 & 73.30 & 85.23 & 40.90 & 64.15 & \underline{93.80} & \underline{97.01} & 81.21 & 90.52  \\
			\cline{1-14} \rowcolor[rgb]{0.9,0.9,0.9}
			 AIFAD = 27.307 & Extreme-Dynamic & \textbf{72.63} & \textbf{85.79} & 0.0 & 0.0 & 0.0 & 0.0 & 1.6 & 10.22 & \underline{51.00} & \underline{72.14} & 15.55 & 39.18 \\
			\noalign{\hrule height 1pt}
	\end{tabular}}
\end{table*}
The results in Tab.~\ref{Tab:DVTr} show that detection performance consistently deteriorates as background Dynamic Variation increases, confirming that stronger scene dynamics substantially increase detection difficulty. Despite this degradation, the proposed method achieves the best performance across all subsets and maintains a clear advantage under highly dynamic conditions. In particular, it remains effective even in the extreme-dynamic setting, where most competing methods fail dramatically. These results demonstrate that explicitly modeling globally coherent background motions effectively suppresses background-induced interference and yields more robust target localization under complex dynamic environments, further supporting our hypothesis that false-alarm accumulation is closely associated with dominant coherent scene dynamics.

\subsection{Limitations and Future Work}
\subsubsection{Limitation} 
\begin{figure}[!t]
	\centering
	\includegraphics[width=\linewidth]{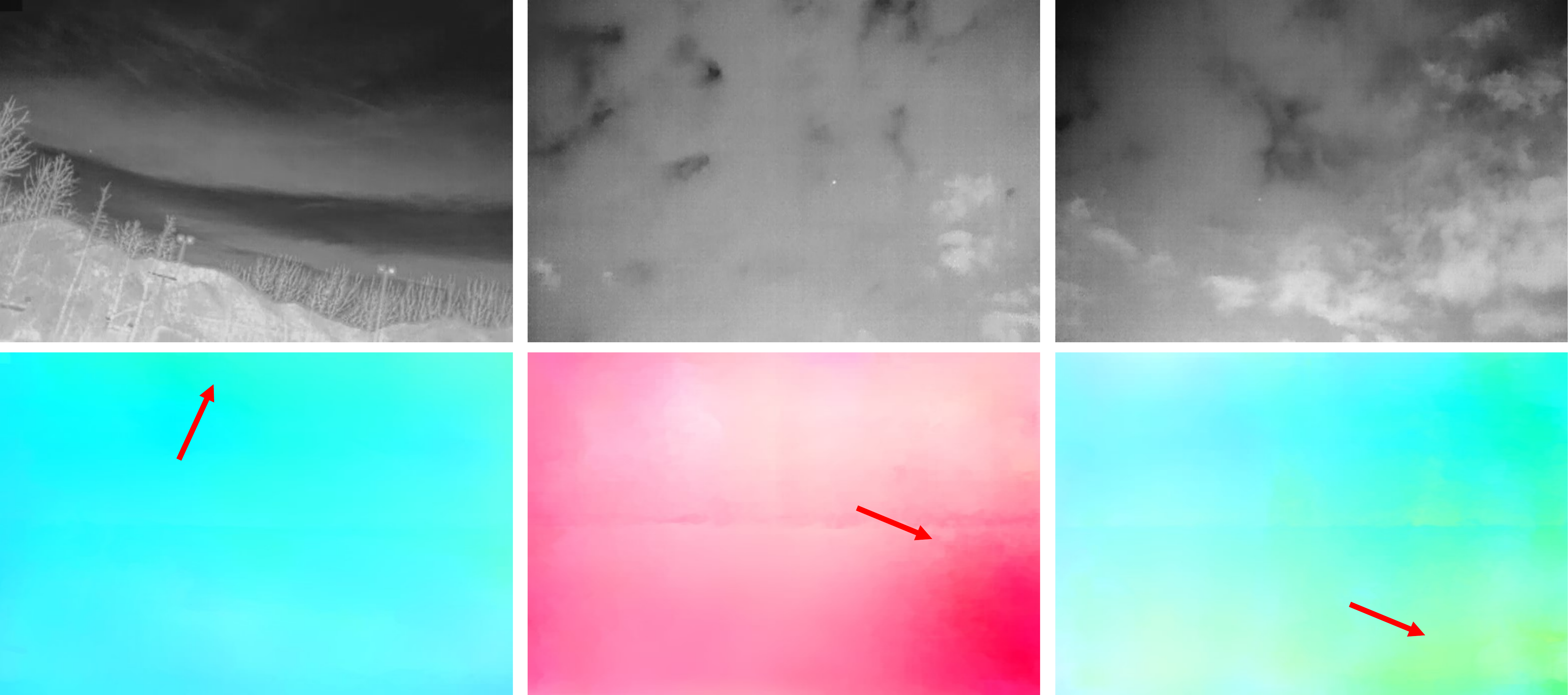}
	\caption{Typical failure cases. Regions entering or leaving the field of view violate the visibility assumption and introduce erroneous reconstruction supervision, which may further cause spatially propagated abnormal optical flow responses.}
	\label{fig:Failed}
\end{figure}

Although the proposed structure-preserving optical-flow adaptation strategy effectively facilitates coherent motion learning, it relies on the visibility assumption underlying self-supervised photometric reconstruction. In practical infrared scenarios, platform motion and dynamic scene changes may cause regions to enter or leave the field of view, violating photometric consistency and introducing unreliable supervision, as illustrated in Fig. \ref{fig:Failed}. Consequently, erroneous optical flow estimation may corrupt coherent motion modeling and further degrade false-alarm suppression and target enhancement.
To investigate this effect, we analyze the relationship between the coherence of the learned global motion and detection performance. As reported in Tab. \ref{Tab:motion_map_corr}, the proposed coherent-motion representation generally satisfies $\nabla \boldsymbol{\mathcal{M}}_t^{g}\approx 0$, validating the global consistency assumption introduced in Sec. \ref{Section:ProblemFormulation}. Moreover, detection performance consistently deteriorates as $mean(\nabla \boldsymbol{\mathcal{M}}_t^{g})$ increases, indicating that deviations from coherent motion are closely associated with inaccurate motion estimation and performance degradation.
\begin{table}[!t]
	\centering
	\caption{Correlation between global motion quality and detection performance.}
	\label{Tab:motion_map_corr}
	\resizebox{0.48\textwidth}{!}{
		\begin{tabular}{c|ccccc}
			\noalign{\hrule height 1pt}
			
			$mean(\nabla \boldsymbol{\mathcal{M}}_t^{g})$ & 0.0100  & 0.0068 & 0.0056 & 0.0040 & 0.0037 \\
			\cline{1-6}
			
			mAP$_{50}(\%)$ & 51.30  & 83.88 & 92.99 & 97.74 & 99.90 \\
			
			\noalign{\hrule height 1pt}
	\end{tabular}}
\end{table}
\subsubsection{Future Work}A promising direction is to exploit auxiliary platform-motion information, which is often available in practical infrared systems, to provide coarse global motion priors and improve the robustness of correspondence learning under severe viewpoint changes and occlusions. However, such metadata alone cannot characterize the localized motion anomalies of tiny targets or false alarms, making fine-grained motion representation learning still indispensable. Since existing public infrared benchmarks rarely provide synchronized platform-motion annotations, future work will focus on constructing large-scale datasets with platform-motion metadata and developing more robust motion learning paradigms that jointly leverage explicit motion priors and localized anomaly cues.

\section{Conclusion}\label{Section:Conclusion}
In this paper, we proposed a motion-decoupled representation learning framework for infrared small target detection under complex dynamic backgrounds. Unlike existing methods that implicitly model highly coupled motion patterns, the proposed framework explicitly decouples background-consistent global motions and target-related local motion anomalies through complementary explicit and implicit motion branches.
To improve motion perception in infrared scenarios, we further introduced a structure-preserving optical-flow adaptation strategy with self-supervised temporal correspondence learning, where phase-aware enhancement and band-pass filtering were incorporated to improve the robustness of motion representation learning against thermal noise and low-texture degradation. Extensive experiments on multiple benchmarks demonstrate the superiority of the proposed framework over existing state-of-the-art methods. We hope this work can provide useful insights into motion-aware representation learning for infrared vision tasks.

\bibliographystyle{IEEEtran}
\bibliography{reference}

\newpage

\vfill

\end{document}